\documentclass[]{bytedance}

\usepackage[toc,page,header]{appendix}
\usepackage{colortbl}   %
\usepackage{amsmath,amssymb}
\usepackage{pifont}
\usepackage{xcolor}
\usepackage{wrapfig}

\usepackage{minitoc}
\usepackage{float}

\title{ Let ViT Speak: Generative Language-Image Pre-training }

\author[1,2,*]{Yan Fang}
\author[2,3,*]{Mengcheng Lan}
\author[2, \dagger]{Zilong Huang}
\author[2]{Weixian Lei}
\author[2]{Yunqing Zhao}
\author[2]{Yujie Zhong}
\author[2]{Yingchen Yu}
\author[2]{Qi She}
\author[1]{Yao Zhao}
\author[1,4, \dagger]{Yunchao Wei}

\affiliation[1]{Beijing Jiaotong University}
\affiliation[2]{ByteDance}
\affiliation[3]{Nanyang Technological University}
\affiliation[4]{\mbox{Beijing Academy of Artificial Intelligence}}

\contribution[*]{Equal contribution}
\contribution[\dagger]{Corresponding authors}

\abstract{
In this paper, we present \textbf{Gen}erative \textbf{L}anguage-\textbf{I}mage \textbf{P}re-training (GenLIP), a simple generative pretraining framework for Vision Transformers (ViTs) designed for multimodal large language models (MLLMs). To better align vision encoders with the autoregressive nature of LLMs, GenLIP trains a ViT to predict language tokens directly from visual tokens using a standard language modeling objective, without contrastive batch construction or an additional text decoder. This design offers three key advantages: (1) \textbf{Simplicity}: a single transformer jointly models visual and textual tokens; (2) \textbf{Scalability}: it scales effectively with both data and model size; and (3) \textbf{Performance}: it achieves competitive or superior results across diverse multimodal benchmarks. Using only one-fifth as many seen samples as SigLIP2 (8B vs. 40B), GenLIP matches or surpasses strong baselines. After continued pretraining on multi-resolution images at native aspect ratios, GenLIP further improves on detail-sensitive tasks such as OCR and chart understanding, making it a strong foundation for vision encoders in MLLMs.
}

\date{September 9, 2026}
\correspondence{
\email{zilong.huang2020@gmail.com}, and \email{yunchao.wei@bjtu.edu.cn}
}

\checkdata[Code and Models]{\href{https://yanfangcs.github.io/vitspeak}{vitspeak}}

\begin{document}
\maketitle

{\renewcommand{\thefootnote}{}\footnotetext[3]{This work was completed while Yan Fang and Mengcheng Lan were interns at ByteDance.}}

\section{Introduction}
\label{sec:intro}

Multimodal Large Language Models (MLLMs) have emerged as a transformative paradigm in artificial intelligence, demonstrating remarkable capabilities in understanding and reasoning across vision and language modalities~\cite{liu2023llava,zhu2024minigpt,sun2024generative, qwenvl, chen2024internvl}.
The prevailing architecture of MLLMs comprises three core components: a vision encoder for processing visual information~\cite{alexander2021vit, radford2021clip, cherti2023openclip, zhai2023siglip}, a connector for bridging modalities, and a large language model (LLM) as the reasoning engine~\cite{achiam2023gpt, touvron2023llama, bai2023qwen, qwen2.5}.
Among these components, the vision encoder serves as the \textit{perceptual foundation}, responsible for extracting meaningful visual representations that can be effectively consumed by the downstream LLM.
Consequently, the quality and design of this vision encoder fundamentally determine the upper bound of an MLLM's visual understanding capability.
As a result, large-scale Vision-Language Pre-training (VLP) on billions of image-text corpora has become the dominant approach for developing strong vision encoders.

\begin{figure}[t]
    \centering
    \includegraphics[width=1.0\linewidth]{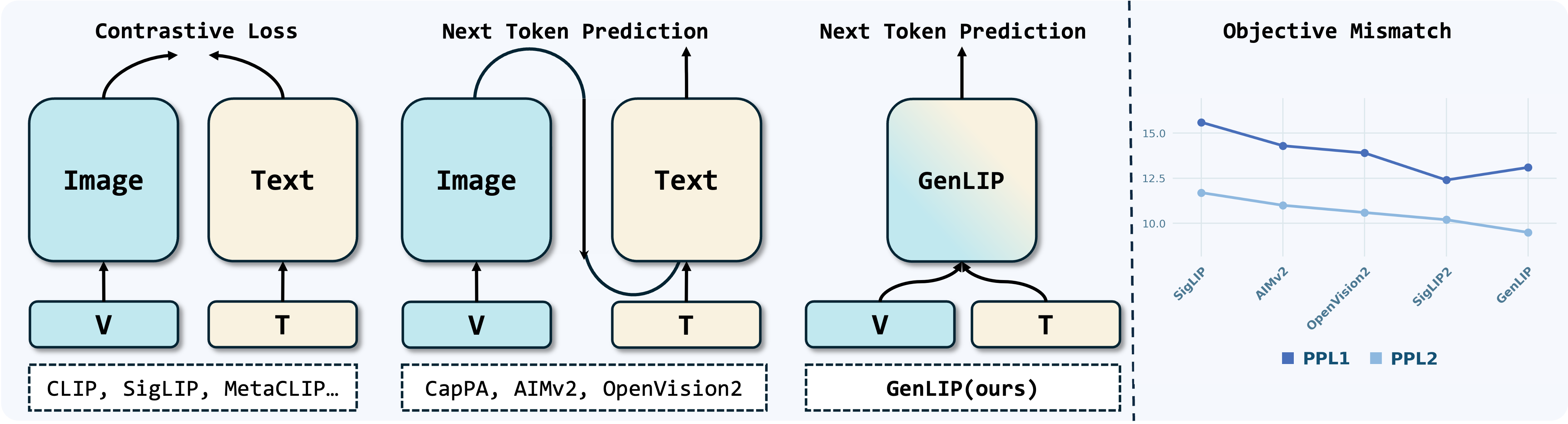}
    \caption{\textbf{Overview of GenLIP and compatibility analysis.}
    \textbf{Left:} Compared with prior vision-language pretraining methods that rely on dual-encoder structures or additional text modules, GenLIP adopts a simplified architecture that directly trains visual tokens with next-token prediction. We use ``V'' and ``T'' to denote visual and textual inputs, respectively.
    \textbf{Right:} We connect different vision encoders to an LLM through a projector and measure perplexity when tuning only the projector (PPL1) or tuning both the projector and LLM (PPL2). Lower perplexity indicates better compatibility with the LLM's generative objective.}
    \label{fig:compare}
\end{figure}

Contrastive learning based VLP methods, exemplified by CLIP~\cite{radford2021clip} and SigLIP~\cite{zhai2023siglip}, are among the most widely adopted vision encoders in MLLMs~\cite{beyer2024paligemma,team2025kimivl,team2026kimi-k2.5}.
These methods typically employ a dual-encoder architecture that encodes each modality separately and align them using a contrastive objective.
However, contrastive pretraining may not be fully aligned with the generative training paradigm of MLLMs: while contrastive learning encourages discriminative image-text alignment, MLLMs are optimized through next-token prediction.
Consistent with this mismatch, the diagnostic evaluation in Figure~\ref{fig:compare} shows that vision features from generative objectives tend to yield lower perplexity after being attached to an LLM, suggesting better compatibility with the LLM's generation.

Another stream of works focuses on generative pretraining, such as CapPa~\cite{tschannen2023cappa}, AIMv2~\cite{fini2025aimv2}, and OpenVision2~\cite{liu2025openvision2}.
These methods typically couple a vision encoder with a text decoder and train the resulting model with an autoregressive language modeling objective.
In this setup, the vision encoder is optimized indirectly through gradients that pass through the text decoder.
Related hybrid designs, such as CoCa~\cite{yu2022coca} and SigLIP2~\cite{tschannen2025siglip2}, further introduce a text encoder to combine contrastive and generative objectives.
While these approaches narrow the gap, their architectural redundancy and indirect optimization complicate training and can limit efficiency when the goal is to learn a scalable vision encoder for MLLMs.

To unleash the full potential of generative vision-language pretraining, we advocate for a simple design philosophy: remove unnecessary modules and train the vision backbone as directly as possible.
Following this principle, we propose a simplified framework for generative vision-language pretraining: \textbf{Gen}erative \textbf{L}anguage-\textbf{I}mage \textbf{P}retraining (GenLIP), a simple yet scalable framework that departs from the complex designs of prior VLP methods. Instead of introducing novel architectural components, our core insight is elegantly simple: \textit{let the Vision Transformer (ViT) speak directly}--requiring no contrastive batch construction and no additional text module.

Instead of indirectly optimizing the vision encoder through additional text components, GenLIP directly trains a ViT to predict language tokens that describe visual content using only a standard autoregressive language modeling objective.
This simple generative formulation aligns the vision encoder more naturally with the way MLLMs operate, while also simplifying the architecture and improving scalability.

GenLIP's design philosophy offers three compelling advantages:
\textbf{(1) Simplicity:} GenLIP uses a single vision backbone and a standard autoregressive objective, without contrastive losses or additional text modules;
\textbf{(2) Scalability:} it scales effectively with both data and model size, yielding consistent gains in our experiments; and
\textbf{(3) Performance:} it achieves competitive or superior results as a vision encoder for MLLMs, with particularly strong performance on optical character recognition (OCR) tasks.
Across extensive experiments, GenLIP matches or outperforms strong baselines pretrained on much larger corpora while using only 8B pretraining samples, and its second-stage native-aspect-ratio adaptation further improves downstream performance.

In summary, GenLIP provides a direct and efficient formulation of generative vision-language pretraining.
Our results suggest that a simpler and better-aligned pretraining paradigm can serve as a strong foundation for future MLLMs.
We believe these findings chart a more direct, efficient, and scalable course for developing powerful vision-language models.

\section{Related Work}
\label{sec:related}
The convergence of computer vision and natural language processing has been driven by large-scale vision-language pretraining, which aims to learn robust, generalizable multimodal representations from massive image-text corpora.
Typical VLP methods can be grouped into three categories based on architectural design and training objectives: dual-encoder contrastive pretraining, encoder-decoder generative pretraining, and simplified single-transformer pretraining.

\textbf{Dual-Encoder Contrastive Pretraining.}
A broad line of research has investigated Contrastive Language-Image Pretraining.
CLIP-style architectures~\cite{radford2021clip,cherti2023openclip,jia2021scaling,cherti2023reproducible,zhai2023siglip,xu2023demystifying} are fundamentally based on a dual-encoder (two-tower) design, which learns to align image and text representations within a shared embedding space using an InfoNCE or similar contrastive objective.
Subsequent works improve alignment by leveraging high-quality image-text pairs~\cite{fan2023improving,zheng2024dreamlip,lai2024veclip,yang2023alip,gadre2023datacomp,li2025openvision,chuang2025metaclip2} or dense region-level captions~\cite{zhang2022glipv2,li2024densefusion,li2025denseworld} for fine-grained representation learning.
While effective for discriminative tasks such as classification and retrieval, contrastive pretraining primarily focuses on global alignment and does not facilitate deep cross-modal interaction.

\textbf{Encoder-Decoder Generative Pretraining.}
To enable richer cross-modal reasoning, recent works~\cite{wang2021simvlm,alayrac2022flamingo,wang2022git,fini2025aimv2,liu2025openvision2} adopt generative pretraining, typically cascading a vision encoder with a text decoder.
For example, Aimv2~\cite{fini2025aimv2} couples a vision encoder with a multimodal decoder that autoregressively generates raw image patches and text tokens,
whereas CapPa~\cite{tschannen2023cappa}, GIT~\cite{wang2022git} and OpenVision 2~\cite{liu2025openvision2} stack a text decoder on top of the image encoder and pretrain the model using only a captioning loss.
Most recently, some studies~\cite{li2021align,li2022blip,yu2022coca,li2025openvision,tschannen2025siglip2,liu2024clips} form hybrid pretraining schemes that combine a contrastive dual-encoder for image-text alignment with a generative decoder for captioning.

\noindent\textbf{Discussion.} Despite their success, existing methods often rely on multiple towers or multiple optimization objectives, which increases model complexity and limits efficiency.
Moreover, alignment is often performed at later stages rather than within the image encoder itself, which can constrain early cross-modal interactions.
In contrast, we propose a simple generative vision-language pretraining framework with a simplified architecture and training objective--a single transformer and a single language modeling objective.

\textbf{Single-Transformer Pretraining.}
Recently, some works also explored vision-language pretraining under a simplified single-Transformer architecture with different objectives.
Among them, SuperClass~\cite{huang2024classification} proposes vision transformer pretraining with a single Transformer tower using token-level classification targets.
VL-BEiT~\cite{bao2022vl} and OneR~\cite{jang2023unifying} aim to unify vision-language representation learning within a single-tower Transformer, but still rely on multiple objectives.
Beyond vision transformer pretraining, several recent efforts~\cite{diao2024eve,chen2024solo,team2024chameleon,diao2025evev2,lei2025sail,diao2025pixels} aim to build native MLLMs with a single transformer and a single language modeling objective.

\textbf{Discussion.} In particular, GenLIP is architecturally close to SAIL~\cite{lei2025sail}, as both use a single transformer with a language modeling objective. However, SAIL focuses on building a native MLLM with a simplified architecture based on pretrained LLMs, whereas GenLIP is designed to pretrain a scalable vision encoder from scratch to better serve modular MLLMs~\cite{chen2024internvl,li2024llavaonevision,bai2025qwen3vl}. This distinct goal also leads to different design choices. Moreover, our controlled comparison suggests that SAIL's LLM initialization is not necessarily beneficial when the goal is to obtain a strong standalone vision encoder, further distinguishing GenLIP from these works.
\section{Approach}
\label{sec:methods}
This section details GenLIP, our simple implementation for generative vision-language pretraining.
We first introduce the core designs of our approach, including the model architecture, data representation, and training objective, all designed for our simplified generative vision-language pretraining.
We then provide pretraining details, including pretraining datasets and training schedule.

\begin{figure*}
    \centering
    \includegraphics[width=1.0\linewidth]{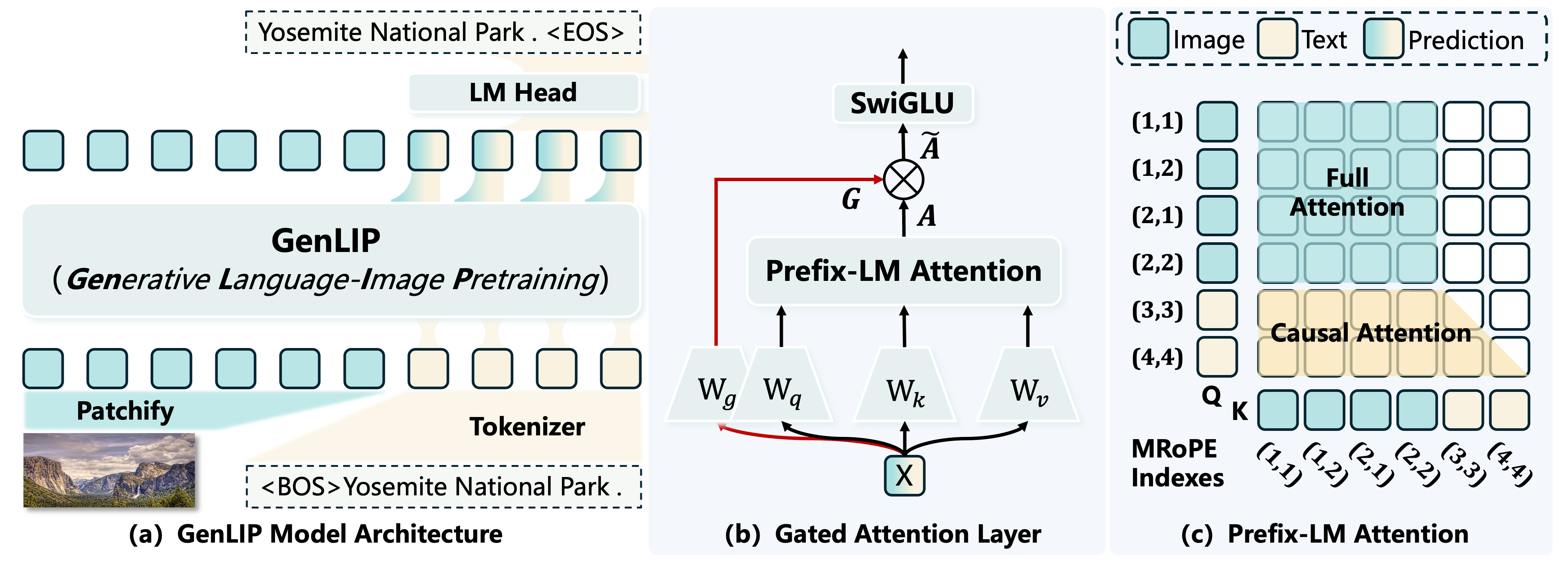}
    \caption{An overview of the GenLIP framework for simple generative vision-language pretraining.
        \textbf{(a) GenLIP Model Architecture}: a single Transformer architecture processes a concatenated visual-prefix sequence. The next token prediction is performed exclusively on text tokens via a language modeling head.
        \textbf{(b) Gated Attention Layer}: the basic layer of GenLIP. The red line in the figure shows the forward path of the gating signal, which is element-wise multiplied with the attention output to control information flow.
        \textbf{(c) Prefix-LM Attention Mechanism}: image tokens attend bidirectionally, while text tokens attend causally. Multimodal Rotary Position Encoding (MRoPE) injects position information into the query (Q) and key (K) vectors.
        }
    \label{fig:model_arch}
\end{figure*}

\subsection{GenLIP Framework}
Instead of introducing novel architectural components, GenLIP is built upon a simple unified modeling paradigm for vision encoder pretraining.
Specifically, we build GenLIP with a simple transformer architecture in the spirit of \textit{letting the ViT speak directly}, analogous to how LLMs generate text.
We keep the design simple, introducing only minimal but necessary modifications for improving visual representations.

\paragraph{\textbf{Data Format.}}
All pretraining data for GenLIP is structured as image-text samples, denoted as $\{(I_i, T_i)\}_{i=1}^N$, where each image $I_i$ is associated with its caption $T_i$.
Each image $I_i$ is partitioned into a sequence of non-overlapping patches $\{v_0,v_1,...,v_M\}$ using a convolutional patch embedding layer, as in standard ViT models.
The corresponding text $T_i$ is tokenized into a sequence of subword tokens $\{t_0,t_1,...,t_L\}$ using an off-the-shelf text tokenizer (Qwen3~\cite{yang2025qwen3}).
The resulting image patch embeddings and text token embeddings are concatenated into a single sequence, with the image embeddings preceding the text embeddings.
The final input sequence $S$ for a given pair $(I_i, T_i)$ is:
\begin{equation}
S = [v_0, \dots, v_M, t_0, \dots, t_L].
\end{equation}

\paragraph{\textbf{Architecture.}}
The architecture of GenLIP is centered around a unified Transformer encoder that processes a concatenated sequence of image and text tokens.
As illustrated in Figure~\ref{fig:model_arch}, the model consists of four components: modality-specific embedding layers, a unified Transformer with a prefix-LM attention implementation, a Layer Normalization (LN) layer, and finally a language modeling (LM) head for token prediction.

To enable effective cross-modal interactions and unified modeling of the concatenated visual-prefix multimodal sequence, we make two small but crucial modifications to a standard Transformer.
(i) To better encode the position information in a concatenated visual-prefix multimodal sequence, we use multimodal rotary position encoding (MRoPE)~\cite{wang2024qwen2vl} and discard the absolute position embeddings for image patches.
(ii) We replace the basic full attention with prefix-LM attention~\cite{raffel2020exploring} in all transformer blocks, where image tokens attend bidirectionally and text tokens attend causally.
Based on the above two modifications, we directly apply the GenLIP architecture to process the unified multimodal sequence, without additional modality-specific designs in the network architecture.

\paragraph{\textbf{Objective.}}
GenLIP adopts a single standard autoregressive language modeling objective, applied exclusively to the textual part of the sequence.
The model is trained to predict the next text token conditioned on the preceding image tokens and text tokens, thereby directly modeling the conditional probability distribution $P(T|I)$.
The objective is to minimize the negative log-likelihood of the text sequence:
\begin{equation}
\mathcal{L}_{\text{LM}} = - \sum_{k=0}^{L} \log P(t_k | \{v_j\}_{j=0}^M, \{t_i\}_{i=0}^{k-1}; \theta)
\label{eq:objective}
\end{equation}
where $\theta$ denotes the model parameters to be optimized, and $P(t_k | \{v_j\}_{j=0}^M, \{t_i\}_{i=0}^{k-1})$ is the predicted probability of the $k$-th text token conditioned on all preceding visual and textual tokens.

\paragraph{\textbf{Using GenLIP as a Vision Encoder.}} 
When employing GenLIP as a visual encoder, we extract vision features from the output of the LN layer following the last Transformer block and feed them into a 2-layer MLP projector to align them with the LLM's input space.
In this process, the language modules of GenLIP (the tokenizer and LM head) are discarded because no text inputs are used, while all other components are retained. The Prefix-LM attention mechanism reduces to standard full attention when GenLIP is used as a vision encoder.

\begin{figure*}[b]
    \centering
    \includegraphics[width=1.0\linewidth]{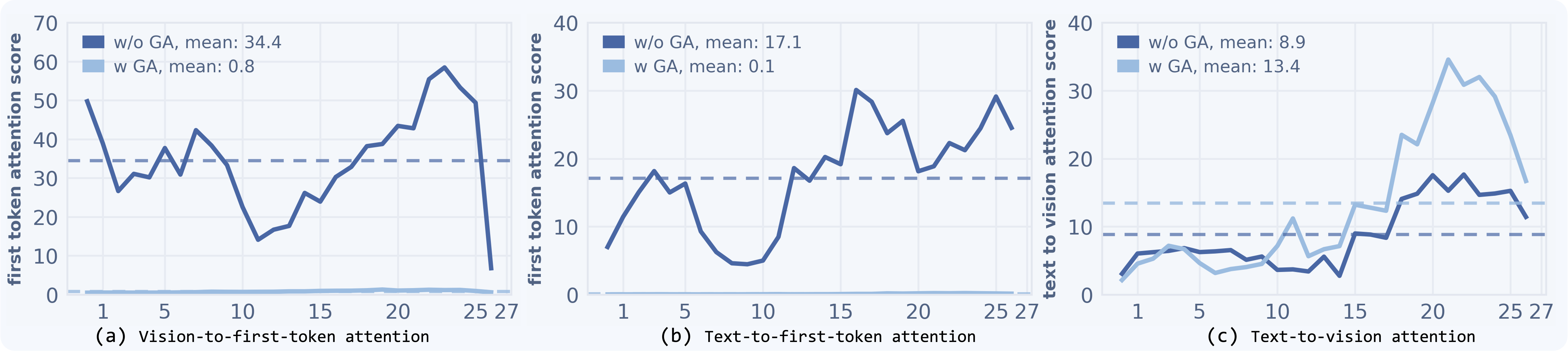}
    \caption{\textbf{Layer-wise attention allocation in controlled So/16 models}. We analyze the attention allocation of different source tokens across 27 layers. The X-axis represents the layer index. We report the attention mass from (a) vision tokens to the first sequence token, (b) generated text tokens to the first sequence token, and (c) generated text tokens to vision tokens. Dashed lines denote the layer-averaged attention scores. GA effectively reduces the first-token attention sink in both vision and text streams, while strengthening text-to-vision attention in deeper layers.}
    \vspace{-12pt}
    \label{fig:att_sink}
\end{figure*}

\subsection{Gated Attention}
While the above unified architecture is effective for generative vision-language pretraining, we observe a notable side effect: attention becomes overly concentrated on the first token of the input sequence, a phenomenon known as the \textit{attention sink}.
This issue is particularly pronounced in our mixed-modality setting, as shown in Figure~\ref{fig:att_sink}.
Under full attention, certain visual tokens can freely aggregate global information from all patches, effectively becoming image-level summaries.
Since text tokens only access visual information through causal attention over this shared visual prefix, the model learns a shortcut: compressing visual information into a few sink tokens for efficient language prediction, at the cost of degrading spatial diversity in visual representations.
Consistent with findings in~\cite{qiu2025gated}, this leads to (i) obvious loss spikes during pretraining, and (ii) attention distributions where the first token absorbs most of the attention mass, reducing the effective utilization of visual tokens.
As a result, the pretrained ViT exhibits substantially degraded discriminative performance, such as in ImageNet attentive probing, and shows unstable scaling behavior--both undesirable for our target usage as a vision encoder for MLLMs.

Inspired by~\cite{qiu2025gated}, we introduce a gated attention mechanism to regulate information flow in the mixed-modality modeling space.
Given input hidden states $X \in \mathbb{R}^{n \times d}$ for a Transformer block, we compute a standard attention output $A = \mathrm{Attn}(X)$ and apply a per-head input-dependent gate:
\begin{equation}
G = \sigma(X W_g + b_g), \quad \widetilde{A} = G \odot A,
\end{equation}
where $\sigma(\cdot)$ is the sigmoid function, $W_g$ and $b_g$ are learnable parameters, and $\odot$ denotes element-wise multiplication. 
The gated attention output $\widetilde{A}$ is then used in the standard residual pathway.
By modulating attention outputs on a per-token basis, the gate prevents text tokens from collapsing their attention onto a small subset of visual tokens and encourages the model to leverage spatially distributed visual features.
In practice, gated attention alleviates loss spikes, accelerates convergence, and stabilizes scaling behavior.

\subsection{Pretraining Details}
Our pretraining comprises two stages with different datasets and resolutions, progressing from fixed low-resolution inputs to diverse resolutions and native aspect ratios.
This setup allows the model to learn foundational visual and linguistic representations while keeping the overall computational cost manageable.

\paragraph{\textbf{Fixed Resolution Pretraining.}}
We first pretrain GenLIP on Recap-DataComp-1B~\cite{xianhang2024recap}, a large-scale dataset of 1 billion unique image-text samples collected from the web.
During this stage, we use the fixed 224$\times$224 resolution images to reduce computational cost while learning strong foundational visual representations.
We train GenLIP for a total of 8 billion samples in this stage, corresponding to 8 epochs over the dataset.

\paragraph{\textbf{Diverse Resolution Adaptation.}} We further fine-tune fixed-resolution pretrained GenLIP on public open-source caption datasets,
namely the caption subset of Infinity-MM (stage1)~\cite{gu2024infinity}, BLIP3o-Long-Caption~\cite{chen2025blip3}, and CapRL~\cite{xing2025caprl}, containing 39 million image-text samples with long captions and higher-resolution images.
Different from higher resolution adaptation in previous works~\cite{oquab2024dinov2,tschannen2025siglip2}, we process images in their native aspect ratios, and resize them to keep the number of vision tokens within $[16, 1024]$.
In this adaptation stage, we train GenLIP for only 1 epoch over the datasets.
This stage helps the model adapt to variable-resolution inputs and learn finer-grained visual representations from dense text descriptions, which is important for downstream tasks that require detailed visual understanding and precise image-text grounding.

\paragraph{\textbf{Regularization.}} We apply two regularization techniques during GenLIP pretraining to effectively train deeper networks: layer scale and drop path. These techniques mainly stabilize training and prevent divergence when training deeper models, although we find that they have limited impact on the final GenLIP performance.

\begin{table*}[!htbp]
    \caption{Overview of GenLIP configurations and pretraining setup. \textbf{Left:} model configurations. \textbf{Right:} two-stage pretraining details.}
    \label{tab:model_config}
    \label{tab:pretrain_details}
    \vspace{-10pt}
    \centering
    \begin{minipage}[t]{0.43\textwidth}
        \centering
        \scriptsize{Model configurations.}\\
        \small
        \resizebox{\linewidth}{!}{
        \begin{tabular}{l c c c c c}
        \toprule
            \textbf{Model}       & \textbf{Params}  & \textbf{Layers}  & \textbf{Dims} & \textbf{Heads} & \textbf{FFN-w} \\
        \midrule
            GenLIP-L    & 0.3B  & 24   & 1024    & 16   & 2752    \\
            GenLIP-So   & 0.4B  & 27   & 1152    & 16   & 3072    \\
            GenLIP-g    & 1.1B  & 40   & 1536    & 24   & 4096    \\
        \bottomrule
        \end{tabular}
        }
    \end{minipage}
    \hfill
    \begin{minipage}[t]{0.55\textwidth}
        \centering
        \scriptsize{Two-stage pretraining details.}\\
        \small
        \resizebox{\linewidth}{!}{
        \begin{tabular}{l | l | c c c c}
        \toprule
            \textbf{Stage}       & \textbf{Dataset} & \textbf{Size} & \textbf{Resolution} & \textbf{Patches} & \textbf{Samples}   \\
        \midrule
            S1 & Recap-DataComp-1B  & 1.0B   & 224     & 196     & 8.0B       \\
            \hline
            \multirow{3}{*}{S2} & Blip3o-Long-Cap & 27M  & \multirow{3}{*}{AnyRes} & \multirow{3}{*}{[16,1024]} & \multirow{3}{*}{39M}  \\
            & Infinity-MM-Stage1 & 10M & & & \\
            & CapRL              & 2M  & & & \\
        \bottomrule
        \end{tabular}
        }
    \end{minipage}
\end{table*}

\paragraph{\textbf{Pretraining Implementation}.}
\begin{wraptable}{r}{0.5\linewidth}
    \vspace{-12pt}
    \caption{Hyperparameters in GenLIP pretraining. ``Batch Size'' denotes the estimated global sample batch size.}
    \label{tab:hyperparameters}
    \small
    \centering
    \vspace{-12pt}
    \begin{tabular}{l ccc}
        \toprule
        Config & L/16 & So/16 & g/16 \\
        \midrule
        Optimizer & \multicolumn{3}{c}{PyTorch AdamW} \\
        Momentum  & \multicolumn{3}{c}{$\beta_1 = 0.9$, $\beta_2 = 0.95$} \\
        Peak LR        & \multicolumn{3}{c}{1e-3}\\
        Min LR    & \multicolumn{3}{c}{1e-6} \\
        LR Decay  & \multicolumn{3}{c}{cosine decay} \\
        Warmup Ratio & 0.007 & 0.007 & 0.02 \\
        Gradient Clipping & \multicolumn{3}{c}{1.0} \\
        Max Packing Length & \multicolumn{3}{c}{16384} \\
        Batch Size & 32K & 32K & 48K \\
        \midrule
        Layer Scale & \multicolumn{3}{c}{0.1} \\
        Drop Path Ratio & 0.1 & 0.1 & 0.2 \\
        Vocab Size  & \multicolumn{3}{c}{151936} \\
        RoPE Theta  & \multicolumn{3}{c}{10000} \\
        \bottomrule
    \end{tabular}
    \vspace{-24pt}
\end{wraptable}
Table~\ref{tab:hyperparameters} summarizes the main pretraining hyperparameters for the two stages.
We use a packing strategy to pack samples of variable lengths into long sequences with a maximum length of $16{,}384$. The packed sequences are then batchified to improve training efficiency and hardware utilization.
On top of this packing strategy, we implement exact per-sample Prefix-LM attention using PyTorch FlexAttention, which supports variable sequence lengths and arbitrary attention masks.
For image preprocessing, we use only resize and crop operations without additional augmentations on Recap-DataComp-1B~\cite{xianhang2024recap}.

There are three major differences in the second stage:
(i) the global batch size is reduced from 32K or 48K to 3.6K because the average sample length increases from 270 tokens to about 1200 tokens;
(ii) the peak learning rate is reduced to $1e\!-4$; and
(iii) images are processed at their native aspect ratios.
All other training settings are kept the same as in the first stage.

\subsection{Discussion}
Rather than introducing novel architectural components, GenLIP pursues the simplest possible paradigm for vision encoder pretraining, enabling seamless integration into MLLMs. Here, we summarize the key differences between GenLIP and prior works.

\textbf{Differences from previous generative works.}
GenLIP differs from previous generative vision-language pretraining works~\cite{bao2022vl,tschannen2023cappa,fini2025aimv2,liu2025openvision2} in several key aspects:
(i) Compared with VL-BEIT~\cite{bao2022vl} and AIMv2~\cite{fini2025aimv2}, GenLIP learns from a single standard autoregressive language modeling objective, without masked image modeling or a pixel reconstruction objective;
(ii) Compared with CapPa~\cite{tschannen2023cappa}, AIMv2~\cite{fini2025aimv2}, and OpenVision2~\cite{liu2025openvision2}, GenLIP discards the additional text decoder and leads to a simplified modeling paradigm with a single unified transformer.

\textbf{Differences from previous single Transformer pretraining works.}
GenLIP also differs from previous single Transformer pretraining works~\cite{lei2025sail,diao2025pixels}:
(i) GenLIP focuses on pretraining a scalable vision encoder for modular MLLMs, rather than native MLLMs;
(ii) GenLIP is pretrained from scratch on caption datasets, while SAIL~\cite{lei2025sail} and NEO~\cite{diao2025pixels} are trained by leveraging pretrained LLMs and large-scale instruction-tuning data;
(iii) GenLIP improves the attention implementation with a gated mechanism that better fits visual modeling when the model is used as a vision encoder.

\section{Experiments}
\label{sec:experiments}

To comprehensively evaluate the visual features learned by GenLIP, we first describe the experimental setup and benchmark suite, then report frozen-representation and standard LLaVA-NeXT evaluations on multimodal understanding tasks.
We next analyze scalability with respect to both pretraining data and model size.
We further conduct controlled comparisons and ablations on the pretraining method, SAIL-style architecture and initialization, gated attention, and native-aspect-ratio adaptation.
Finally, we evaluate the discriminative transferability of the learned features and provide qualitative ``Let ViT Speak'' analyses through direct caption generation and patch-semantics readout.

\subsection{Setup}
\label{subsec:4_1eval}

\subsubsection{Baselines}
We compare our method, GenLIP, against a suite of representative vision-language pre-training models under multimodal understanding benchmarks.
This includes contrastive methods such as CLIP~\cite{radford2021clip}, SigLIP~\cite{zhai2023siglip}, and SigLIP2~\cite{tschannen2025siglip2}, as well as generative approaches like AIMv2~\cite{fini2025aimv2} and OpenVision2~\cite{liu2025openvision2}.
For a fair comparison in the frozen evaluation, all vision encoders are configured to produce the same number of visual tokens before the projector: L/14 encoders are evaluated at $336\times336$, while L/16, So/16, and g/16 encoders are evaluated at $384\times384$, yielding a $24\times24$ patch grid for each model.
We use strong publicly available model variants for our baselines, such as ViT-L/14 for CLIP and AIMv2, and ViT-So/16 for SigLIP2.
These methods are pretrained on substantially larger training corpora (12.0B--40.0B image-text pairs) than GenLIP.

\subsubsection{Experimental Setup}
Following Cambrian~\cite{tong2024cambrian}, we mainly adopt frozen visual representation evaluation, where the vision encoder is kept frozen and the language model is fine-tuned on downstream tasks.
This protocol directly measures the quality of visual features learned by different VLP methods without the confounding effect of further fine-tuning the vision encoder.
Based on the LLaVA-NeXT framework~\cite{liu2024llavanext}, we replace the original vision encoder with one pretrained by GenLIP or each baseline method, and then fine-tune the language model on an instruction tuning dataset.
To better unleash the potential of the vision encoders, we replace the original 780K instruction-tuning set with the comprehensive LLaVA OneVision~\cite{li2024llavaonevision} dataset, which contains more than 3 million supervised fine-tuning (SFT) samples.
We consider two LLM backbones of different sizes in our implementation, Qwen2.5-1.5B-Instruct and Qwen2.5-7B-Instruct~\cite{qwen2025qwen25technicalreport}, in place of the original LLM in LLaVA-NeXT. 
In our implementation, we adopt a standard 2-layer MLP as the projector. For baselines, vision features are extracted from the final block of the ViT and subsequently fed into the LLM via the projector. For GenLIP, we extract vision features from the last LN layer based model architecture.

\subsubsection{Evaluation Benchmarks}
To provide a comprehensive evaluation, we assess our method and all baselines across a diverse set of multimodal understanding benchmarks.
These benchmarks are grouped into three categories to probe distinct capabilities: document understanding and optical character recognition (Doc\&OCR), general visual understanding (General VQA), and image captioning (Caption). All evaluations are conducted using the LMMS-Eval toolkit~\cite{zhang2025lmms}.

\textbf{Document and OCR.} This category evaluates the model's ability to recognize and interpret text within images, a critical skill for document analysis and scene text understanding.
Following mainstream MLLMs~\cite{bai2025qwen3vl,li2024llavaonevision}, we focus on a wide range of classic benchmarks, including ChartQA~\cite{masry2022chartqa}, OCRBench~\cite{liu2024ocrbench}, InfoVQA~\cite{mathew2022infographicvqa}, AI2D~\cite{kembhavi2016ai2d}, TextVQA~\cite{singh2019textvqa}, DocVQA~\cite{mathew2021docvqa} and SEED-Bench-2-Plus~\cite{li2024seedbench2}.

\textbf{General Visual Understanding.} This group of tasks assesses the model's broader capabilities in comprehending and reasoning about visual content.
We employ four widely-used benchmarks, including MME~\cite{fu2023mme}, GQA~\cite{hudson2019gqa}, VQAv2~\cite{goyal2017vqav2}, and ScienceQA~\cite{lu2022scienceqa} for general VQA. 

\textbf{Image Captioning.} To measure the model's ability to generate descriptive text from images, we evaluate on NoCaps~\cite{agrawal2019nocaps}, COCO~\cite{mao2016generation}, and TextCaps~\cite{sidorov2020textcaps} for evaluation. Performance is reported using the CIDEr metric.

For a holistic comparison, we report an overall average score across all 14 benchmarks (ALL AVG), computed as the mean of the per-benchmark scores. In particular, we divide MME-P scores by 20 to map their original $[0, 2000]$ range to $[0, 100]$ before averaging, ensuring comparability. Besides these benchmarks, we also extend this evaluation suite to the Cambrian-1 style and provide results in following.

\definecolor{lightcyan}{HTML}{eaf3f3}
\definecolor{lightblue}{HTML}{F4F7FC}
\newcommand{\B}[1]{\textbf{#1}}

\subsection{Main Results}
We provide all frozen visual representation evaluation results on multimodal understanding benchmarks in Table~\ref{tab:Qwen2_5-1_5B} and Table~\ref{tab:Qwen2_5-7B}.
Besides, we also provide results under the standard unfrozen LLaVA-NeXT evaluation setting in Table~\ref{table:LLaVA-NeXT-Eval}.

\FloatBarrier

\begin{table}[!htbp]
  \caption{Frozen visual representation evaluation under LLaVA-NeXT-Qwen2.5-1.5B. We test GenLIP models across three scales against baseline methods. The benchmarks are grouped into Doc\&OCR, General VQA, and Caption tasks. ``Arch'' stands for ``Model Architecture'', while ``Data'' denotes ``Pretraining Data Scale''. ``OpenVision2'' is abbreviated as ``OVision2''.}
  \label{tab:Qwen2_5-1_5B}
  \centering
  \resizebox{\linewidth}{!}{
  \begin{tabular}{lcc | ccccccc  cccc ccc c}
    \toprule
    \multirow{2}{*}{\B{Model}} & \multirow{2}{*}{\B{Arch}} & \multirow{2}{*}{\B{Data}} & \multicolumn{7}{c}{\B{Doc\&OCR}} & \multicolumn{4}{c}{\B{General VQA}} & \multicolumn{3}{c}{\B{Caption}} \\
    \cmidrule(lr){4-10} \cmidrule(lr){11-14} \cmidrule(lr){15-17}
    & &
    & \rotatebox{90}{\scriptsize{ChartQA}}
    & \rotatebox{90}{\scriptsize{OCR-B}}
    & \rotatebox{90}{\scriptsize{DocVQA}}
    & \rotatebox{90}{\scriptsize{TextVQA}}
    & \rotatebox{90}{\scriptsize{AI2D}}
    & \rotatebox{90}{\scriptsize{InfoVQA}}
    & \rotatebox{90}{\scriptsize{SEED-2}}
    & \rotatebox{90}{\scriptsize{VQAv2}}
    & \rotatebox{90}{\scriptsize{GQA}}
    & \rotatebox{90}{\scriptsize{SQA}}
    & \rotatebox{90}{\scriptsize{MME-P}}
    & \rotatebox{90}{\scriptsize{NoCaps}}
    & \rotatebox{90}{\scriptsize{COCO}}
    & \rotatebox{90}{\scriptsize{TextCaps}}
    & \multirow[t]{2}{*}{\rotatebox{90}{ALL AVG}}
    \\
    \midrule
    CLIP~\cite{radford2021clip}               & L/14  & 12.8B & 24.8 & 23.7 & 38.9 & 43.9 & 64.5 & 30.2 & 47.8 & 46.1 & 39.8 & 75.3 & 1218 & 55.5 & 72.5 & 117.9 & 53.1 \\
    AIMv2~\cite{fini2025aimv2}                & L/14  & 12.0B & 26.3 & 25.2 & 37.7 & 47.2 & 64.2 & 29.3 & 47.3 & \textbf{48.1} & \B{43.9} & 76.2 & 1157 & 80.1 & 73.6 & 122.4 & 55.7  \\
    OVision2~\cite{liu2025openvision2}     & L/16  & 12.8B & 30.7 & 45.6 & 43.3 & 49.2 & 65.6 & 28.1 & 47.8 & 44.0 & 42.7 & 75.5 & 1230 & \B{84.3} & \B{76.3} & 127.4 & 58.7 \\
    SigLIP~\cite{zhai2023siglip}              & L/16  & 40.0B & 30.2 & 41.0 & 47.3 & 36.0 & 66.4 & 27.8 & 47.9 & 41.3 & 41.7 & 76.7 & 1203 & 84.0 & 76.1 & 120.7 & 56.9 \\
    SigLIP2~\cite{tschannen2025siglip2}       & L/16  & 40.0B & 33.4 & 45.7 & 45.1 & 50.3 & \textbf{66.7} & 28.2 & 45.7 & 43.1 & 42.6 & \textbf{76.9} & 1165 & 82.9 & 74.6 & 127.8 & 58.7  \\
    \rowcolor{lightcyan}
    GenLIP                                    & L/16  & 8.0B  & \textbf{41.2} & \textbf{51.1} & \textbf{51.1} & \textbf{53.6} & 66.6 & \textbf{30.7} & \textbf{51.1} & 44.4 & 41.5 & 76.1 & \textbf{1258} & 82.6 & 76.0 & \B{131.4} & \B{61.5}  \\
    \midrule
    SigLIP2~\cite{tschannen2025siglip2}       & So/16 & 40.0B & 35.2 & 47.2 & 46.4 & 53.3 & 67.0 & 28.0 & 50.3  & \textbf{46.5} & 43.5 & \textbf{77.1} & \textbf{1220} & 84.3 & 77.1 & \B{131.5} & 60.6 \\
    \rowcolor{lightcyan}
    GenLIP                                    & So/16 & 8.0B  & \textbf{40.8} & \textbf{51.5} & \textbf{51.9} & \textbf{55.2} & \textbf{67.2} & \textbf{31.9} & \textbf{52.3} & \textbf{46.5} & \textbf{44.0} & 76.0 & 1215 & \B{87.5} & \B{81.5} & 129.5 & \B{62.6}  \\
    \midrule
    SigLIP2~\cite{tschannen2025siglip2}       & g/16 & 40.0B & 35.3 & 47.3 & 47.6 & 54.7 & 66.7 & 29.7 & 49.6 & \textbf{50.1} & 45.2 & 76.2 & \textbf{1284} & 84.4 & 76.2 & 134.5 & 61.5 \\
    \rowcolor{lightcyan}
    GenLIP                                    & g/16 & 8.0B  & \textbf{45.0} & \textbf{55.6} & \textbf{57.0} & \textbf{59.0} & \textbf{68.9} & \textbf{33.9} & \textbf{53.3} & 49.1 & \textbf{45.5} & \textbf{77.5} & 1256 & \B{88.3} & \B{82.0} & \B{135.4} & \B{65.2} \\
    \bottomrule
  \end{tabular}
}
\end{table}

\subsubsection{Frozen Feature Analysis}
As presented in Table~\ref{tab:Qwen2_5-1_5B}, GenLIP demonstrates strong performance across three model scales.
Despite using fewer pretraining pairs, GenLIP achieves consistent gains over all baselines, including the 40B-pair pretrained SigLIP2. Under the Qwen2.5-1.5B setting, GenLIP improves the overall average (ALL AVG) over SigLIP2 by $2.8$, $2.0$, and $3.7$ points at the L/16, So/16, and g/16 scales, respectively.
The gains are especially pronounced on Doc\&OCR benchmarks, which demand fine-grained document understanding and text-centric visual reasoning. Averaging over the seven Doc\&OCR tasks in Table~\ref{tab:Qwen2_5-1_5B}, GenLIP achieves $49.3$, $50.1$, and $53.2$ at L/16, So/16, and g/16, outperforming SigLIP2 by $4.3$, $3.3$, and $5.9$ points, respectively.

This advantage remains under a larger LLM. As shown in Table~\ref{tab:Qwen2_5-7B}, scaling the LLM to Qwen2.5-7B yields consistent trends with the Qwen2.5-1.5B setting. Under this setting, GenLIP outperforms SigLIP2 by $2.4$ and $4.7$ points on average score at the So/16 and g/16 scales, respectively.
Similar to the Qwen2.5-1.5B setting, GenLIP consistently performs best on Doc\&OCR benchmarks, highlighting its strong visual-text alignment.

Across both frozen settings, GenLIP not only surpasses contrastive VLMs such as CLIP~\cite{radford2021clip} and SigLIP~\cite{zhai2023siglip}, but also outperforms prior encoder-decoder generative VLMs, including AIMv2~\cite{fini2025aimv2} and OpenVision2~\cite{liu2025openvision2}. These generative baselines use an additional text decoder for language modeling, and OpenVision2 is further pretrained on the stronger Recap-DataComp-1B v2 corpus with a longer training schedule. Overall, the results suggest that GenLIP's simple architecture and objective can yield stronger visual representations with improved data efficiency.

We also observe that GenLIP scales favorably with model size, while SigLIP2 shows comparatively smaller gains when scaling up. These results support two hypotheses: (i) simplifying both the architecture and the objective can enable more efficient scaling; and (ii) larger model capacity helps GenLIP learn both broad visual knowledge and fine-grained alignment for multimodal understanding.

\begin{table}[!htbp]
  \caption{Frozen visual representation evaluation under LLaVA-NeXT-Qwen2.5-7B. Except for the LLM size, all settings are the same as those used in LLaVA-NeXT-Qwen2.5-1.5B.}
  \label{tab:Qwen2_5-7B}
  \centering
  \resizebox{\linewidth}{!}{
  \begin{tabular}{lcc | ccccccc cccc ccc c}
    \toprule
    \multirow{2}{*}{\B{Model}} & \multirow{2}{*}{\B{Arch}} & \multirow{2}{*}{\B{Data}} & \multicolumn{7}{c}{\B{Doc\&OCR}} & \multicolumn{4}{c}{\B{General VQA}} & \multicolumn{3}{c}{\B{Caption}} \\
    \cmidrule(lr){4-10} \cmidrule(lr){11-14} \cmidrule(lr){15-17}
    & &
    & \rotatebox{90}{\scriptsize{ChartQA}}
    & \rotatebox{90}{\scriptsize{OCR-B}}
    & \rotatebox{90}{\scriptsize{DocVQA}}
    & \rotatebox{90}{\scriptsize{TextVQA}}
    & \rotatebox{90}{\scriptsize{AI2D}}
    & \rotatebox{90}{\scriptsize{InfoVQA}}
    & \rotatebox{90}{\scriptsize{SEED-2}}
    & \rotatebox{90}{\scriptsize{VQAv2}}
    & \rotatebox{90}{\scriptsize{GQA}}
    & \rotatebox{90}{\scriptsize{SQA}}
    & \rotatebox{90}{\scriptsize{MME-P}}
    & \rotatebox{90}{\scriptsize{NoCaps}}
    & \rotatebox{90}{\scriptsize{COCO}}
    & \rotatebox{90}{\scriptsize{TextCaps}}
    & \multirow[t]{2}{*}{\rotatebox{90}{ALL AVG}}
    \\
    \midrule
    CLIP~\cite{radford2021clip}               & L/14  & 12.8B & 36.6 & 29.6 & 48.4 & 52.6 & 76.3 & \B{39.0} & 55.1  & 49.4 & 39.6 & 85.2 & 1316 & 63.1 & 54.4 & 127.9 & 58.8 \\
    AIMv2~\cite{fini2025aimv2}                & L/14  & 12.0B & 36.8 & 30.9 & 46.6 & 54.5 & 76.9 & 37.5 & 55.1  & 44.0 & 37.9 & 85.2 & 1240 & 66.5 & 55.9 & 130.5 & 58.6 \\
    OVision2~\cite{liu2025openvision2}     & L/16  & 12.8B & 42.5 & 49.9 & 49.5 & 58.8 & 78.4 & 33.8 & 53.8  & \B{60.0} & 47.2 & 85.9 & \B{1325} & 79.4 & 69.6 & 133.8 & 64.9 \\
    SigLIP~\cite{zhai2023siglip}              & L/16  & 40.0B & 41.7 & 45.7 & 50.5 & 56.0 & 79.3 & 34.8 & 55.8  & 57.8 & 46.2 & \B{86.7} & 1275 & \B{81.5} & \B{72.0} & 131.1 & 64.5 \\
    \rowcolor{lightcyan}
    GenLIP                                    & L/16  & 8.0B  & \B{52.7} & \B{59.2} & \B{61.7} & \B{62.9} & \B{80.4} & 38.8 & \B{59.0}  & 56.4 & \B{51.3} & 85.4 & 1320 & 81.1 & 71.3 & \B{139.4} & \B{69.0} \\
    \midrule
    SigLIP2~\cite{tschannen2025siglip2}       & So/16 & 40.0B & 46.6 & 55.6 & 56.3 & 63.5 & \B{81.3} & 37.2 & 56.4 & \B{64.5} & 52.2 & \B{87.1} & 1422 & \B{84.1} & \B{76.4} & 139.3 & 69.4 \\
    \rowcolor{lightcyan}
    GenLIP                                    & So/16 & 8.0B  & \B{55.3} & \B{63.5} & \B{66.3} & \B{65.7} & 81.0 & \B{41.4} & \B{60.8}  & 60.5 & \B{52.4} & 86.4 & \B{1424} & 83.1 & 74.8 & \B{142.1} & \B{71.8} \\
    \midrule
    SigLIP2~\cite{tschannen2025siglip2}       & g/16  & 40.0B & 47.2 & 55.6 & 56.3 & 63.5 & \B{81.0} & 36.4 & 56.4 & 62.7 & 49.3 & \B{87.7} & 1422 & 82.0 & 72.3 & 142.7 & 68.9 \\
    \rowcolor{lightcyan}
    GenLIP                                    & g/16  & 8.0B  & \B{57.1} & \B{65.9} & \B{69.0} & \B{66.8} & \B{81.0} & \B{43.6} & \B{61.1} & \B{64.4} & \B{54.5} & 87.0 & \B{1483} & \B{85.0} & \B{75.5} & \B{144.8} & \B{73.6} \\
    \bottomrule
  \end{tabular}
  }
\end{table}

\subsubsection{Cambrian-1 Evaluation Suite}

Based on the frozen evaluation in Table~\ref{tab:Qwen2_5-1_5B} and Table~\ref{tab:Qwen2_5-7B}, we also organize our evaluation results in a Cambrian-1-style~\cite{tong2024cambrian} benchmark suite by adding benchmarks such as MMMU~\cite{yue2023mmmu}, MathVista~\cite{lu2024mathvista}, MMVP~\cite{tong2024eyes}, and RealWorldQA\cite{realworldqa2024}. Table~\ref{tab:cambrian_qwen2_5_1_5B} shows detailed results under LLaVA-NeXT-Qwen2.5-1.5B settings. On this extended evaluation suite, GenLIP still shows its performance advantage against strong baselines on all 3 model sizes. Results under LLaVA-NeXT-Qwen2.5-7B can be found in appendix.

\begin{table}[!htbp]
  \caption{Cambrian-1-style frozen visual representation evaluation under LLaVA-NeXT-Qwen2.5-1.5B. The benchmarks are grouped into General VQA, Knowledge, OCR \& Chart, and Vision-Centric tasks. ``MMB'', ``SEED-I'', ``MathV'', ``RWQA'', ``CV-2D'', and ``CV-3D'' denote MMBench~\cite{liu2024mmbench}, SEED-Image~\cite{li2024seednbench}, MathVista~\cite{lu2024mathvista}, RealWorldQA\cite{realworldqa2024}, CV-Bench2D~\cite{tong2024cambrian}, and CV-Bench3D~\cite{tong2024cambrian}, respectively. We report results on MMBench$\_$en subset for MMBench, and MathVista testmini subset with format score for MathVista.}
  \label{tab:cambrian_qwen2_5_1_5B}
  \centering
  \resizebox{\linewidth}{!}{
  \begin{tabular}{lcc | cccc  cccc  cccc  cccc  c}
    \toprule
    \multirow{2}{*}{\B{Model}} & \multirow{2}{*}{\B{Arch}} & \multirow{2}{*}{\B{Data}} & \multicolumn{4}{c}{\B{General VQA}} & \multicolumn{4}{c}{\B{Knowledge}}  & \multicolumn{4}{c}{\B{OCR \& Chart}} & \multicolumn{4}{c}{\B{Vision-Centric}} \\
    \cmidrule(lr){4-7} \cmidrule(lr){8-11} \cmidrule(lr){12-15} \cmidrule(lr){16-19}
    & &
    & \rotatebox{90}{\scriptsize{MME-P}}
    & \rotatebox{90}{\scriptsize{MMB}}
    & \rotatebox{90}{\scriptsize{SEED-I}}
    & \rotatebox{90}{\scriptsize{GQA}}
    & \rotatebox{90}{\scriptsize{SQA}}
    & \rotatebox{90}{\scriptsize{MMMU}}
    & \rotatebox{90}{\scriptsize{MathV}}
    & \rotatebox{90}{\scriptsize{AI2D}}
    & \rotatebox{90}{\scriptsize{ChartQA}}
    & \rotatebox{90}{\scriptsize{OCR-B}}
    & \rotatebox{90}{\scriptsize{TextVQA}}
    & \rotatebox{90}{\scriptsize{DocVQA}}
    & \rotatebox{90}{\scriptsize{MMVP}}
    & \rotatebox{90}{\scriptsize{RWQA}}
    & \rotatebox{90}{\scriptsize{CV-2D}}
    & \rotatebox{90}{\scriptsize{CV-3D}}
    & \multirow[t]{2}{*}{\rotatebox{90}{AVG}} \\
    \midrule
    CLIP~\cite{radford2021clip} & L/14 & 12.8B & 1218 & 63.7 & 64.1 & 39.8 & 75.3 & 37.9 & 36.1 & 64.5 & 24.8 & 23.7 & 43.9 & 38.9 & 31.3 & 44.4 & 49.6 & 48.3 & 46.7 \\
    AIMv2~\cite{fini2025aimv2} & L/14 & 12.0B & 1157 & 66.2 & 65.6 & \B{43.9} & 76.2 & \B{39.1} & 38.6 & 64.2 & 26.3 & 25.2 & 47.2 & 37.7 & 34.0 & 50.1 & 52.7 & \B{54.8} & 48.7 \\
    OVision2~\cite{liu2025openvision2} & L/16 & 12.8B & 1230 & 66.4 & 65.5 & 42.7 & 75.5 & 38.6 & 35.9 & 65.6 & 30.7 & 45.6 & 49.2 & 43.3 & 30.7 & \B{52.0} & 53.6 & 52.8 & 50.6 \\
    SigLIP~\cite{zhai2023siglip} & L/16 & 40.0B & 1203 & 67.3 & 66.2 & 41.7 & 76.7 & 38.0 & 37.4 & 66.4 & 30.2 & 41.0 & 36.0 & 47.3 & 33.3 & 47.3 & 53.4 & 51.4 & 49.6 \\
    SigLIP2~\cite{tschannen2025siglip2} & L/16 & 40.0B & 1165 & \B{67.4} & \B{67.0} & 42.6 & \B{76.9} & 36.9 & 39.0 & \B{66.7} & 33.4 & 45.7 & 50.3 & 45.1 & 37.3 & 51.8 & \B{56.2} & 54.3 & 51.8 \\
    \rowcolor{lightcyan}
    GenLIP & L/16 & 8.0B & \B{1258} & 65.8 & \B{67.0} & 41.5 & 76.1 & 36.3 & \B{40.5} & 66.6 & \B{41.2} & \B{51.1} & \B{53.6} & \B{51.1} & \B{38.0} & 49.3 & 55.8 & 54.2 & \B{53.2} \\
    \midrule
    SigLIP2~\cite{tschannen2025siglip2} & So/16 & 40.0B & \B{1220} & \B{66.8} & 66.3 & 43.5 & \B{77.1} & 38.2 & 38.4 & 67.0 & 35.2 & 47.2 & 53.3 & 46.4 & \B{39.3} & 50.3 & 53.5 & \B{52.5} & 52.2 \\
    \rowcolor{lightcyan}
    GenLIP & So/16 & 8.0B & 1215 & 66.1 & \B{67.7} & \B{44.0} & 76.0 & \B{40.8} & \B{38.6} & \B{67.2} & \B{40.8} & \B{51.5} & \B{55.2} & \B{51.9} & 37.3 & \B{52.0} & \B{55.2} & 51.7 & \B{53.5} \\
    \midrule
    SigLIP2~\cite{tschannen2025siglip2} & g/16 & 40.0B & \B{1284} & \B{68.0} & 66.9 & 45.2 & 76.2 & \B{38.9} & 39.9 & 66.7 & 35.3 & 47.3 & 54.7 & 47.6 & 36.7 & 52.3 & 53.5 & 54.2 & 53.0 \\
    \rowcolor{lightcyan}
    GenLIP & g/16 & 8.0B & 1256 & 67.2 & \B{68.7} & \B{45.5} & \B{77.5} & 37.8 & \B{41.0} & \B{68.9} & \B{45.0} & \B{55.6} & \B{59.0} & \B{57.0} & \B{40.0} & \B{55.0} & \B{54.1} & \B{54.8} & \B{55.6} \\
    \bottomrule
  \end{tabular}
  }
\end{table}

\subsubsection{Standard LLaVA-NeXT Evaluation}
We further evaluate GenLIP under the standard LLaVA-NeXT setting following prior work~\cite{yinxie_2025_rice}, where the vision encoder is unfrozen and fine-tuned jointly with the language model during instruction tuning.
As shown in Table~\ref{table:LLaVA-NeXT-Eval}, GenLIP performs strongly under two fixed patch budgets and achieves competitive overall results across both Doc\&OCR and General VQA tasks. GenLIP shows consistent advantages on Doc\&OCR benchmarks.

Taken together, both the frozen and standard evaluations indicate that GenLIP provides strong and consistent performance across diverse multimodal understanding tasks, including Doc\&OCR, General VQA, and captioning.
In particular, GenLIP consistently excels on Doc\&OCR tasks, which demand fine-grained visual recognition and precise visual-text alignment.

Overall, these results indicate that GenLIP, a simple yet effective generative vision-language pretraining method, can learn rich and versatile visual representations for multimodal understanding with high data efficiency.
Compared with more complex alternatives (e.g., SigLIP2 with larger pretraining corpora and more elaborate training recipes), GenLIP exhibits highly competitive and often achieves better downstream performance.
This suggests that simple generative vision-language pretraining is a promising direction for learning strong, scalable visual representations for MLLMs.

\begin{table}[!htbp]
  \caption{Multimodal understanding results under standard LLaVA-NeXT settings. All models are evaluated using identical configurations: the same data and LLM and anyres image processing configuration~\cite{liu2024llavanext}.}
  \label{table:LLaVA-NeXT-Eval}
  \centering
  \resizebox{\linewidth}{!}{
  \begin{tabular}{llcc| cccccc  cccccc c}
    \toprule
    &\multirow{2}{*}{\B{Model}} & \multirow{2}{*}{\B{Arch}} & \multirow{2}{*}{\B{Data}} & \multicolumn{6}{c}{\B{Doc\&OCR}} & \multicolumn{6}{c}{\B{General VQA}}  \\
    \cmidrule(lr){5-10} \cmidrule(lr){11-16}
    \multirow[t]{2}{*}{\rotatebox{90}{\B{Patches}}} & & &
    & \rotatebox{90}{\scriptsize{ChartQA}}
    & \rotatebox{90}{\scriptsize{DocVQA}}
    & \rotatebox{90}{\scriptsize{TextVQA}}
    & \rotatebox{90}{\scriptsize{OCR-B}}
    & \rotatebox{90}{\scriptsize{LiveVQA}}
    & \rotatebox{90}{\scriptsize{AI2D}}
    & \rotatebox{90}{\scriptsize{MMBench}}
    & \rotatebox{90}{\scriptsize{MME-C}}
    & \rotatebox{90}{\scriptsize{MME-P}}
    & \rotatebox{90}{\scriptsize{POPE}}
    & \rotatebox{90}{\scriptsize{RWQA}}
    & \rotatebox{90}{\scriptsize{MMStar}}
    & \multirow[t]{2}{*}{\rotatebox{90}{\B{ALL AVG}}}
    \\
    \midrule
    \multirow{5}{*}{576}
    & CLIP~\cite{radford2021clip}   & L/14  & 12.8B & 75.2 & 66.5 & 62.5 & 52.5 & 47.4 & 73.2 & 74.6 & 48.0 & 75.6 & 88.8 & 63.7 & 49.0  & 64.8 \\
    & MLCD~\cite{anxiang_2024_mlcd}                & L/14  & 12.0B & 76.5 & 67.8 & 61.7 & 53.1 & 48.4 & 77.0 & 76.5 & 54.1 & 79.9 & 88.7 & 61.1 & 51.0 & 66.3 \\
    & AIMv2~\cite{fini2025aimv2}   & L/14  & 12.8B & 77.2 & 72.7 & 65.9 & 57.2 & 47.3 & 75.4 & \B{78.6} & 48.3 & 75.0 & 88.4 & 62.2 & 50.2  & 66.5 \\
    & RICE-ViT~\cite{yinxie_2025_rice}     & L/14  & 13.0B & 79.2 & 72.3 & 65.9 & 57.5 & \B{48.9} & 77.9 & 76.6 & \B{54.6} & \B{80.7} & 88.5 & 63.1 & 51.8 & 68.1 \\
    \rowcolor{lightcyan}
    & GenLIP       & So/16 & 8.0B  & \B{79.3} & \B{75.2} & \B{68.5} & \B{59.7} & 48.4 & \B{78.6} & 77.7 & 48.6 & 78.2 & \B{89.2} & \B{65.9} & \B{53.1} & \B{68.5} \\
    \midrule
    \multirow{4}{*}{729}
    & SigLIP~\cite{zhai2023siglip}                   & So/14  & 40.0B & 76.7 & 69.3 & 64.7 & 55.4 & 48.4 & 76.2 & 77.0 & 46.1 & 79.9 & 88.8 & 63.7 & 47.3 & 66.1\\
    & SigLIP2~\cite{tschannen2025siglip2}           & So/14 & 40.0B & 79.1 & 70.2 & 66.2 & 58.7 & 48.6 & 77.0 & 77.1 & 46.6 & \B{80.4} & 89.3 & 63.4 & 52.8 & 67.5 \\
    & RICE-ViT~\cite{yinxie_2025_rice} & L/14  & 13.0B & 82.6 & 75.1 & 66.2 & 58.8 & 49.5  & 76.5 & 77.6 & 54.1 & 79.0 & 89.1 & 62.9 & 51.2 & 68.6 \\
    \rowcolor{lightcyan}
    & GenLIP                             & So/16 & 8.0B  & \B{83.0} & \B{76.9} & \B{69.6} & \B{64.7} & \B{50.4} & \B{79.1} & \B{78.1} & \B{54.5} & 80.1 & \B{89.4} & \B{65.1} & \B{53.2} & \B{70.3} \\
    \bottomrule
  \end{tabular}
  }
\end{table}

\begin{figure}[t]
    \centering
    \includegraphics[width=\linewidth]{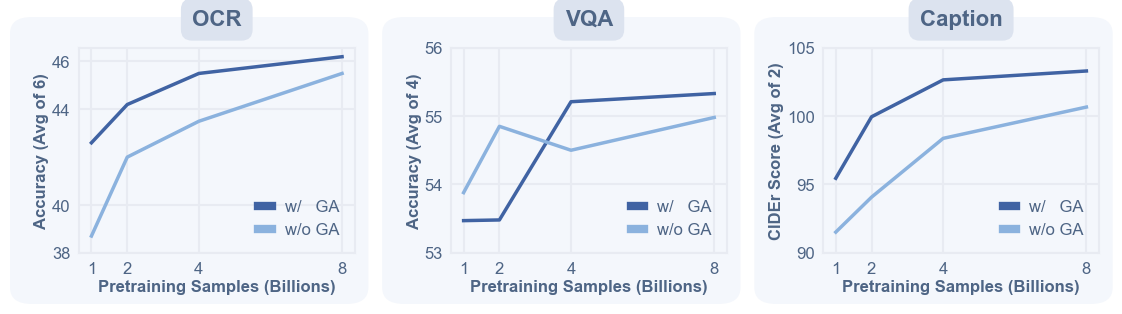}
    \caption{\textbf{Data Scaling Behavior.} Performance on three kinds of tasks as the number of pretraining samples in the first stage is scaled from 1.0B to 8.0B. We report and plot the curve of the average score for Doc\&OCR, VQA, and Caption tasks. The x-axis in each subplot corresponds to the pretraining data scale.}
    \label{fig:data_scaling}
\end{figure}

\subsection{Scalability Analysis}
To investigate the detailed scaling pattern of GenLIP, we discuss both data and model scalability of GenLIP, which are two key factors for VLP pretraining.

\subsubsection{Data Scaling}
We first study data scaling in Fig.~\ref{fig:data_scaling}, where we pretrain GenLIP (with or without gated attention) on Recap-DataComp-1B with different numbers of training samples, ranging from 1.0B to 8.0B.
As the data scale increases from 1.0B to 8.0B, GenLIP shows sustained improvements on multimodal understanding benchmarks.
We observe steeper gains when scaling from 1.0B to 4.0B, while the improvement curve becomes flatter when further scaling to 8.0B.
In particular, the average performance on VQA and caption tasks shows only minor improvements when scaling from 4.0B to 8.0B.
Based on this trend, we use 8.0B samples as the default data scale for GenLIP pretraining in our main results.

\subsubsection{Model Scaling}
We also investigate how GenLIP performance changes with model size by pretraining GenLIP at the L/16, So/16, and g/16 scales.
Besides the final results after diverse resolution adaptation shown in Table~\ref{tab:Qwen2_5-1_5B}, we additionally provide results for models pretrained only with fixed low resolution on Recap-DataComp-1B in Table~\ref{tab:model_scaling}. Across both pretraining stages, GenLIP shows consistent performance gains with increasing model size.
An important observation is that GenLIP-L/16 lags behind GenLIP-So/16 and GenLIP-g/16 only with fixed low-resolution pretraining, while the gap between g/16 and So/16 is relatively small.
This suggests that an appropriate model size is important for GenLIP to learn strong visual representations and better performance on downstream tasks.

\begin{table}[!htbp]
  \caption{Frozen visual representation evaluation of GenLIP pretrained at different model scales across two stages. ``S1'' and ``S2'' denotes the pretraining stage 1 and 2 respectively.}
  \label{tab:model_scaling}
  \centering
  \resizebox{\linewidth}{!}{
  \begin{tabular}{lc | ccccccc  cccc ccc c}
    \toprule
    \multirow{2}{*}{\B{Model}} & \multirow{2}{*}{\B{Arch}} & \multicolumn{7}{c}{\B{Doc\&OCR}} & \multicolumn{4}{c}{\B{General VQA}} & \multicolumn{3}{c}{\B{Caption}} \\
    \cmidrule(lr){3-9} \cmidrule(lr){10-13} \cmidrule(lr){14-16}
    &
    & \rotatebox{90}{\scriptsize{ChartQA}}
    & \rotatebox{90}{\scriptsize{OCR-B}}
    & \rotatebox{90}{\scriptsize{DocVQA}}
    & \rotatebox{90}{\scriptsize{TextVQA}}
    & \rotatebox{90}{\scriptsize{AI2D}}
    & \rotatebox{90}{\scriptsize{InfoVQA}}
    & \rotatebox{90}{\scriptsize{SEED-2}}
    & \rotatebox{90}{\scriptsize{VQAv2}}
    & \rotatebox{90}{\scriptsize{GQA}}
    & \rotatebox{90}{\scriptsize{SQA}}
    & \rotatebox{90}{\scriptsize{MME-P}}
    & \rotatebox{90}{\scriptsize{NoCaps}}
    & \rotatebox{90}{\scriptsize{COCO}}
    & \rotatebox{90}{\scriptsize{TextCaps}}
    & \multirow[t]{2}{*}{\rotatebox{90}{ALL AVG}}
    \\
    \midrule
    SigLIP2       & L/16  & 33.4 & 45.7 & 45.1 & 50.3 & \textbf{66.7} & 28.2 & 45.7 & 43.1 & \textbf{42.6} & \textbf{76.9} & 1165 & \textbf{82.9} & 74.6 & 127.8 & 58.7  \\
    GenLIP-S1      & L/16  & 34.3 & 28.9 & 44.5 & 43.0 & 64.5 & 28.5 & 49.1 & 44.0 & 41.9 & 75.2 & 1136 & 77.3 & 71.0 & 114.1 & 55.2 \\
    GenLIP-S2      & L/16  & \textbf{41.2} & \textbf{51.1} & \textbf{51.1} & \textbf{53.6} & 66.6 & \textbf{30.7} & \textbf{51.1} & \textbf{44.4} & 41.5 & 76.1 & \textbf{1258} & 82.6 & \textbf{76.0} & \B{131.4} & \B{61.5}  \\
    \midrule
    SigLIP2      & So/16  & 35.2 & 47.2 & 46.4 & 53.3 & 67.0 & 28.0 & 50.3  & 46.5 & 43.5 & \textbf{77.1} & \textbf{1220} & 84.3 & 77.1 & \textbf{131.5} & 60.6 \\
    GenLIP-S1       & So/16  & 37.6 & 39.2 & 49.8 & 50.5 & 65.3 & 29.7 & 51.3 & 45.4 & 43.8 & 75.2 & 1157 & 80.9 & 73.6 & 125.7 & 58.9 \\
    GenLIP-S2       & So/16  & \textbf{40.8} & \textbf{51.5} & \textbf{51.9} & \textbf{55.2} & \textbf{67.2} & \textbf{31.9} & \textbf{52.3} & \textbf{46.5} & \textbf{44.0} & 76.0 & 1215 & \B{87.5} & \B{81.5} & 129.5 & \B{62.6}  \\
    \midrule
    SigLIP2       & g/16 & 35.3 & 47.3 & 47.6 & 54.7 & 66.7 & 29.7 & 49.6 & \textbf{50.1} & 45.2 & 76.2 & \textbf{1284} & 84.4 & 76.2 & 134.5 & 61.5 \\
    GenLIP-S1        & g/16  & 34.6 & 42.5 & 53.7 & 53.1 & 65.5 & 29.6 & 51.1 & 45.3 & 43.5 & 75.9 & 1164 & 82.0 & 74.0 & 132.0 & 60.0 \\
    GenLIP-S2        & g/16  & \textbf{45.0} & \textbf{55.6} & \textbf{57.0} & \textbf{59.0} & \textbf{68.9} & \textbf{33.9} & \textbf{53.3} & 49.1 & \textbf{45.5} & \textbf{77.5} & 1256 & \B{88.3} & \B{82.0} & \B{135.4} & \B{65.2}
    \\
    \bottomrule
    \end{tabular}
}
\end{table}

\subsection{Ablations}

\subsubsection{Comparison with Other VLPs}
A key property of GenLIP is data efficiency: as shown above, GenLIP pretrained on 8B pairs can surpass baselines pretrained with substantially larger corpora.
To further validate this property, we conduct a controlled comparison among a contrastive method (SigLIP), an encoder--decoder generative method (OpenVision2), and our GenLIP under the same pretraining data budget.

Specifically, we train SigLIP, OpenVision2, and GenLIP on the same 2.0B samples from Recap-DataComp-1B. For GenLIP, we run only the first pretraining stage and evaluate directly at a $384\times384$ input resolution. For SigLIP and OpenVision2, we pretrain at $224\times224$ and further conduct a short high-resolution adaptation stage at $384\times384$ for 0.2B samples.
For SigLIP, we implement the vanilla sigmoid contrastive loss without additional tricks from SigLIP2~\cite{tschannen2025siglip2}.

We evaluate frozen visual representations of these methods under the same protocol in Table~\ref{table:pretraining-method}. Under the same data budget, GenLIP still achieves strong performance on both Doc\&OCR and General VQA tasks. While GenLIP outperforms the baselines on most benchmarks, it trails OpenVision2 on OCRBench by $6.3$, which is likely related to the absence of high-resolution adaptation in GenLIP under this controlled setting and the known difficulty of dense-text recognition with low-resolution pretraining.

Overall, this controlled comparison supports that our simple generative VLP method can be more data-efficient than both contrastive and prior generative alternatives.

\begin{table}[!htbp]
  \caption{Ablation between different pretraining methods.}
  \label{table:pretraining-method}
  \centering
  \resizebox{\linewidth}{!}{
  \begin{tabular}{lcc| ccccccc cccc ccc c}
    \toprule
    \multirow{2}{*}{\B{Model}} & \multirow{2}{*}{Arch} & \multirow{2}{*}{Data} & \multicolumn{7}{c}{\B{OCR}} & \multicolumn{4}{c}{\B{General VQA}}  & \multicolumn{3}{c}{\B{Caption}}  \\
    \cmidrule(lr){4-10} \cmidrule(lr){11-14} \cmidrule(lr){15-17}
    & & & \rotatebox{90}{\scriptsize{ChartQA}}
    & \rotatebox{90}{\scriptsize{OCR-B}}
    & \rotatebox{90}{\scriptsize{DocVQA}}
    & \rotatebox{90}{\scriptsize{TextVQA}}
    & \rotatebox{90}{\scriptsize{AI2D}}
    & \rotatebox{90}{\scriptsize{InfoVQA}}
    & \rotatebox{90}{\scriptsize{SEED-2}}
    & \rotatebox{90}{\scriptsize{VQAv2}}
    & \rotatebox{90}{\scriptsize{GQA}}
    & \rotatebox{90}{\scriptsize{SQA}}
    & \rotatebox{90}{\scriptsize{MME-P}}
    & \rotatebox{90}{\scriptsize{NoCaps}}
    & \rotatebox{90}{\scriptsize{COCO}}
    & \rotatebox{90}{\scriptsize{TextCaps}}
    & \rotatebox{90}{\scriptsize{ALL AVG}}  \\
    \midrule
    SigLIP  & So/16 & 2.0B  & 26.1 & 36.2 & 38.6 & 44.3 & 64.2 & 25.8 & 46.0 & 42.7 & 39.8 & 75.1 & 1132 & 76.2 & 70.6 & 119.2 & 54.4 \\
    OVision2  & So/16 & 2.0B  & 27.8 & \B{43.2} & 41.2 & 44.7 & 64.1 & 26.8 & 46.3 & 44.2 & 40.3 & 74.8 & \B{1158} & 76.3 & 71.0 & \B{123.3} & 55.9 \\
    \rowcolor{lightcyan}
    GenLIP  & So/16 & 2.0B  & \B{35.0} & 36.9 & \B{46.0} & \B{47.1} & \B{64.9} & \B{29.3} & \B{50.3}  & \B{45.4} & \B{42.0} & \B{75.6} & 1156 & \B{78.7} & \B{71.1} & 121.1 & \B{57.2} \\
    \bottomrule
  \end{tabular}
}
\end{table}

\subsubsection{Comparison with SAIL}
GenLIP shares architectural similarities with SAIL~\cite{lei2025sail}. 
To better understand their differences, we compare SAIL-style Qwen3-0.6B vision encoders with GenLIP under the same 1B-sample pretraining setting in Table~\ref{tab:sail_comparison}. 
Adding gated attention to the Qwen3-initialized SAIL variant brings only a modest overall improvement, while training the same architecture from scratch achieves performance close to GenLIP-So/16. Notably, the parameter count of Qwen3-0.6B is very close to that of GenLIP-So/16, with only minor architectural differences such as group query attention, making it well suited for this comparison. These results suggest that a strong language-model initialization in SAIL is not necessarily beneficial for this vision-encoder pretraining setting; instead, it may bias the model toward language-side shortcuts that are less useful for learning robust visual representations. We further analyze this phenomenon through attention patterns in the appendix.

\newcommand{\cmark}{\textcolor{green!50!black}{\ding{51}}}
\newcommand{\xmark}{\textcolor{red!70!black}{\ding{55}}}

\begin{table}[!htbp]
  \caption{Comparison with SAIL. We follow SAIL and train vision encoders based on the Qwen3-0.6B architecture. We further evaluate gated attention on this architecture and denote this variant with the suffix ``-g''. ``Init'' indicates whether the model is initialized from pretrained Qwen3-0.6B weights. Apart from architecture and initialization, all methods use the same settings and are pretrained with 1B samples from Recap-DataComp-1B.}
  \label{tab:sail_comparison}
  \centering
  \resizebox{\linewidth}{!}{
  \begin{tabular}{lcc| ccccccc cccc ccc c}
    \toprule
    \multirow{2}{*}{\B{Method}} & \multirow{2}{*}{\B{Arch}} & \multirow{2}{*}{\B{Init}} & \multicolumn{7}{c}{\B{Doc\&OCR}} & \multicolumn{4}{c}{\B{General VQA}} & \multicolumn{3}{c}{\B{Caption}} \\
    \cmidrule(lr){4-10} \cmidrule(lr){11-14} \cmidrule(lr){15-17}
    & &
    & \rotatebox{90}{\scriptsize{ChartQA}}
    & \rotatebox{90}{\scriptsize{OCR-B}}
    & \rotatebox{90}{\scriptsize{DocVQA}}
    & \rotatebox{90}{\scriptsize{TextVQA}}
    & \rotatebox{90}{\scriptsize{AI2D}}
    & \rotatebox{90}{\scriptsize{InfoVQA}}
    & \rotatebox{90}{\scriptsize{SEED-2}}
    & \rotatebox{90}{\scriptsize{VQAv2}}
    & \rotatebox{90}{\scriptsize{GQA}}
    & \rotatebox{90}{\scriptsize{SQA}}
    & \rotatebox{90}{\scriptsize{MME-P}}
    & \rotatebox{90}{\scriptsize{NoCaps}}
    & \rotatebox{90}{\scriptsize{COCO}}
    & \rotatebox{90}{\scriptsize{TextCaps}}
    & \multirow[t]{2}{*}{\rotatebox{90}{ALL AVG}}
    \\
    \midrule
    SAIL         & Qwen3-0.6B & \cmark &31.6	&30.2	&39.3	&44.1	&62.5	&25.7	&47.3 &\B{41.1}	&40.6	&\B{74.6}	&1057 &74.2	&71.9	&114.3 &53.6 \\
    SAIL-g       & Qwen3-0.6B & \cmark &30.2	&29.5	&39.4	&43.9	&62.4	&26.2	&47.8 &40.7	&41.3	&74.2	&1141 &81.6 &75.4 &116.9 &54.8 \\
    SAIL-g       & Qwen3-0.6B & \xmark &32.5	&\B{36.0}	&43.6	&44.0	&63.7	&27.4	&48.7 &40.5	&40.6	&\B{74.6}	&1131 &\B{81.7}	&\B{77.2}	&116.4 &56.0 \\
    GenLIP       & ViT-So/16      & \xmark &\B{34.6}	&34.2	&\B{44.1}	&\B{44.7}	&\B{64.6}	&\B{29.1}	&\B{51.1} &40.7	&\B{41.5}	&\B{74.6}	&\B{1144} &79.9	&73.7 &\B{118.4} &\B{56.3} \\
    \bottomrule
  \end{tabular}
  }
\end{table}

\subsubsection{Gated Attention}
In Fig.~\ref{fig:data_scaling}, we plot data scaling curves of GenLIP with and without gated attention, showing consistent advantages of gated attention across data scales.
Gated attention improves data efficiency, especially in the low-data regime, where the variant with gated attention achieves higher performance than the one without.
It also leads to better convergence and improves the final performance by a notable margin.

In Table~\ref{tab:sink}, we compare gated attention with an alternative approaches for mitigating attention sinks: register tokens. 
Register tokens introduce additional tokens before the visual sequence to absorb sink behavior in stead of vision tokens. But we find it is better to also keep register tokens in VLMs for better performance, which is different from previous works~\cite{simeoni2025dinov3,darcet2023registers}.
We also try attention bias method by adding learnable key and value bias terms $k', v' \in \mathbb{R}^d$~\cite{an2025systematic} in attention forwarding but find it challenging to figure it out in our GenLIP pretraining.
Gated attention achieves the better overall average and performs strongest on most Doc\&OCR benchmarks than register method with both 1 or 4 register tokens, suggesting that it provides a more effective and flexible solution in our setting.

\begin{table}[!htbp]
  \caption{Comparison among different solutions to the attention sink problem. We conduct an ablation on GenLIP-So/16 and compare gated attention with register tokens. All methods are trained with 1B samples from Recap-DataComp-1B.}
  \label{tab:sink}
  \centering
  \resizebox{\linewidth}{!}{
  \begin{tabular}{l | cccccccc cccc ccc c}
    \toprule
    \multirow{2}{*}{\B{Method}} & \multicolumn{7}{c}{\B{Doc\&OCR}} & \multicolumn{4}{c}{\B{General VQA}} & \multicolumn{3}{c}{\B{Caption}} \\
    \cmidrule(lr){2-8} \cmidrule(lr){9-12} \cmidrule(lr){13-15}
    & \rotatebox{90}{\scriptsize{ChartQA}}
    & \rotatebox{90}{\scriptsize{OCR-B}}
    & \rotatebox{90}{\scriptsize{DocVQA}}
    & \rotatebox{90}{\scriptsize{TextVQA}}
    & \rotatebox{90}{\scriptsize{AI2D}}
    & \rotatebox{90}{\scriptsize{InfoVQA}}
    & \rotatebox{90}{\scriptsize{SEED-2}}
    & \rotatebox{90}{\scriptsize{VQAv2}}
    & \rotatebox{90}{\scriptsize{GQA}}
    & \rotatebox{90}{\scriptsize{SQA}}
    & \rotatebox{90}{\scriptsize{MME-P}}
    & \rotatebox{90}{\scriptsize{NoCaps}}
    & \rotatebox{90}{\scriptsize{COCO}}
    & \rotatebox{90}{\scriptsize{TextCaps}}
    & \multirow[t]{2}{*}{\rotatebox{90}{ALL AVG}}
    \\
    \midrule
    1 Register        & 30.5	&32.5	&37.2	&40.7	&63.6	&27.4	&47.4 &36.4	&35.6	&75.0	&\B{1154} &74.2	&71.9	&114.3 &53.3 \\
    4 Registers        & 34.5	&30.8	&41.0	&43.8	&63.8	&27.5	&50.2 &39.7 &37.5	&\B{75.5}	&1123 &76.6	&\B{74.1}	&116.9 &55.1 \\
    Gated Attention   & \B{34.6}	&\B{34.2}	&\B{44.1}	&\B{44.7}	&\B{64.6}	&\B{29.1}	&\B{51.1} &\B{40.7}	&\B{41.5}	&74.6 &1144 &\B{79.9}	&73.7 &\B{118.4}    &\B{56.3} \\
    \bottomrule
  \end{tabular}
  }
\end{table}

\subsubsection{Native-Aspect-Ratio Adaptation}
\begin{figure}[!t]
  \centering
  \includegraphics[width=\linewidth]{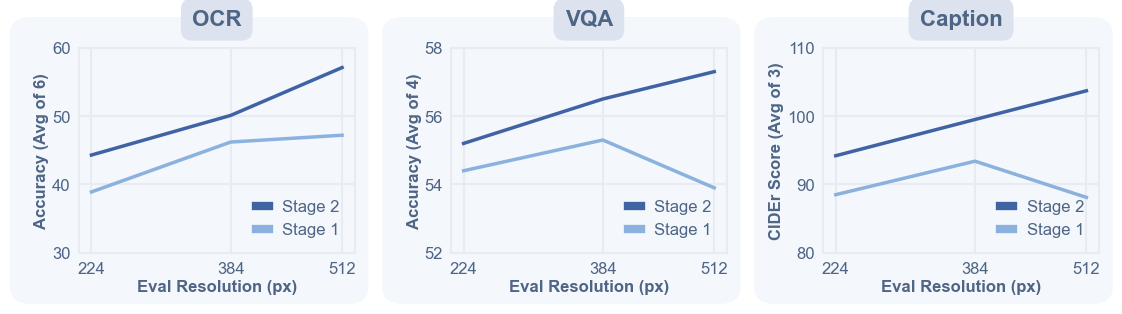}
  \caption{\textbf{Validation of Native Aspect Adaptation.} We evaluate the frozen visual representation of GenLIP-So/16 pretrained after two stages on the same setting as shown in Table~\ref{tab:Qwen2_5-1_5B}. The x-axis corresponds to the input resolution in evaluation, and the y-axis corresponds to the average score on OCR, VQA and Caption tasks, respectively.}
  \label{fig:na-val}
\end{figure}

We evaluate GenLIP pretrained with two stages under different evaluation resolutions, which validates the effectiveness of the native-aspect-ratio adaptation stage.
To test the model's behavior under different input resolutions, we evaluate frozen visual representations of GenLIP (after each stage) across multiple resolutions under the same protocol as in Table~\ref{tab:Qwen2_5-1_5B} (Fig.~\ref{fig:na-val}).

\subsection{Discriminative Ability}
To assess the discriminative quality of GenLIP's visual representations, we adopt the frozen-backbone evaluation protocol from DINOv2~\cite{oquab2024dinov2} and probe the frozen visual features on ImageNet-1K~\cite{deng2009imagenet} for classification and ADE20K~\cite{zhou2017scene} for semantic segmentation. Because GenLIP has no [CLS] token, we use attentive probing on patch features for classification, and use only a linear layer on patch features for semantic segmentation. We extract patch features from last layer of GenLIP, without fusing features from multiple layers. 

\begin{table}[!htbp]
  \caption{Frozen feature evaluation on the ImageNet-1K and ADE20K validation sets. We report top-1 accuracy on ImageNet-1K and mIoU on ADE20K, without test-time augmentation. For GenLIP models, values marked with (S1) use the first-stage model; unmarked values use the second-stage model. ``w/o GA'' denotes the variant without gated attention.}
  \label{tab:frozen_dis}
  \small
  \centering
  \begin{tabular}{lccc}
    \toprule
    Method & Arch &ImageNet-1K & ADE20K \\ 
    \midrule
    CLIP    & L/14  & 85.1 & 39.0 \\
    SigLIP  & So/14 & 86.7 & 40.8 \\
    SigLIP2 & So/16 & 88.9 & 45.4 \\
    \midrule
    GenLIP \scriptsize{w/o GA} & So/16 & 76.2 & - \\
    GenLIP & L/16  & 83.8(S1) / 82.8 & 41.0 \\
    GenLIP & So/16 & 84.3(S1) / 83.1 & 42.8 \\
    GenLIP & g/16  & 85.2(S1) / 83.0 & 44.5 \\
    \bottomrule
  \end{tabular}
\end{table}

As shown in Table~\ref{tab:frozen_dis}, GenLIP learns decent transferable discriminative visual features without explicit visual supervision.
There are two related findings: (i) gated attention effectively alleviates the degraded discriminative representations due to attention sink, and (ii) first-stage ImageNet-1K accuracy and second-stage ADE20K performance improve with model size.
The biggest variant, GenLIP-g/16, achieves $85.2$ top-1 accuracy on ImageNet-1K after Stage 1; after Stage 2, it achieves $83.0$ top-1 accuracy and $44.5$ mIoU on ADE20K. Native-aspect-ratio adaptation improves multimodal understanding but reduces ImageNet-1K accuracy at all three model scales. GenLIP remains below the contrastive baselines on ImageNet-1K, while its ADE20K results exceed those of CLIP and SigLIP but trail SigLIP2, which introduces dense supervision~\cite{tschannen2025siglip2}.
Overall, this result demonstrates our pretraining method delivers competitive visual representations for discriminative tasks with an extremely simple pretraining method.

\subsection{Let ViT Speak}
\label{subsec:4_0vit_speak}

\definecolor{l16s1}{HTML}{528D67}
\definecolor{so16s1}{HTML}{215F9A}
\definecolor{g16s1}{HTML}{7C5FA3}
\definecolor{l16s2}{HTML}{417654}
\definecolor{so16s2}{HTML}{163E64}
\definecolor{g16s2}{HTML}{78548D}
\definecolor{related}{HTML}{3B7D23}
\definecolor{unrelated}{HTML}{C00000}

\subsubsection{Direct Caption Generation}
We begin with a simple but intuitive test of GenLIP's generative ability by asking the model to describe an input image directly.
We evaluate all three model scales on both common-image examples (Figure~\ref{fig:caption-1}) and supplementary OCR-heavy examples reported in the appendix (Figure~\ref{fig:caption-2}). For this test, we use temperature=$1e-6$, top$_p$=1.0, a maximum of 256 new tokens, and no beam search. Generation stops when the model outputs the end-of-sequence token. We use the simple prompt ``Describe the image in details.'' throughout.

\begin{figure}[!htb]
  \centering
  \includegraphics[width=\linewidth]{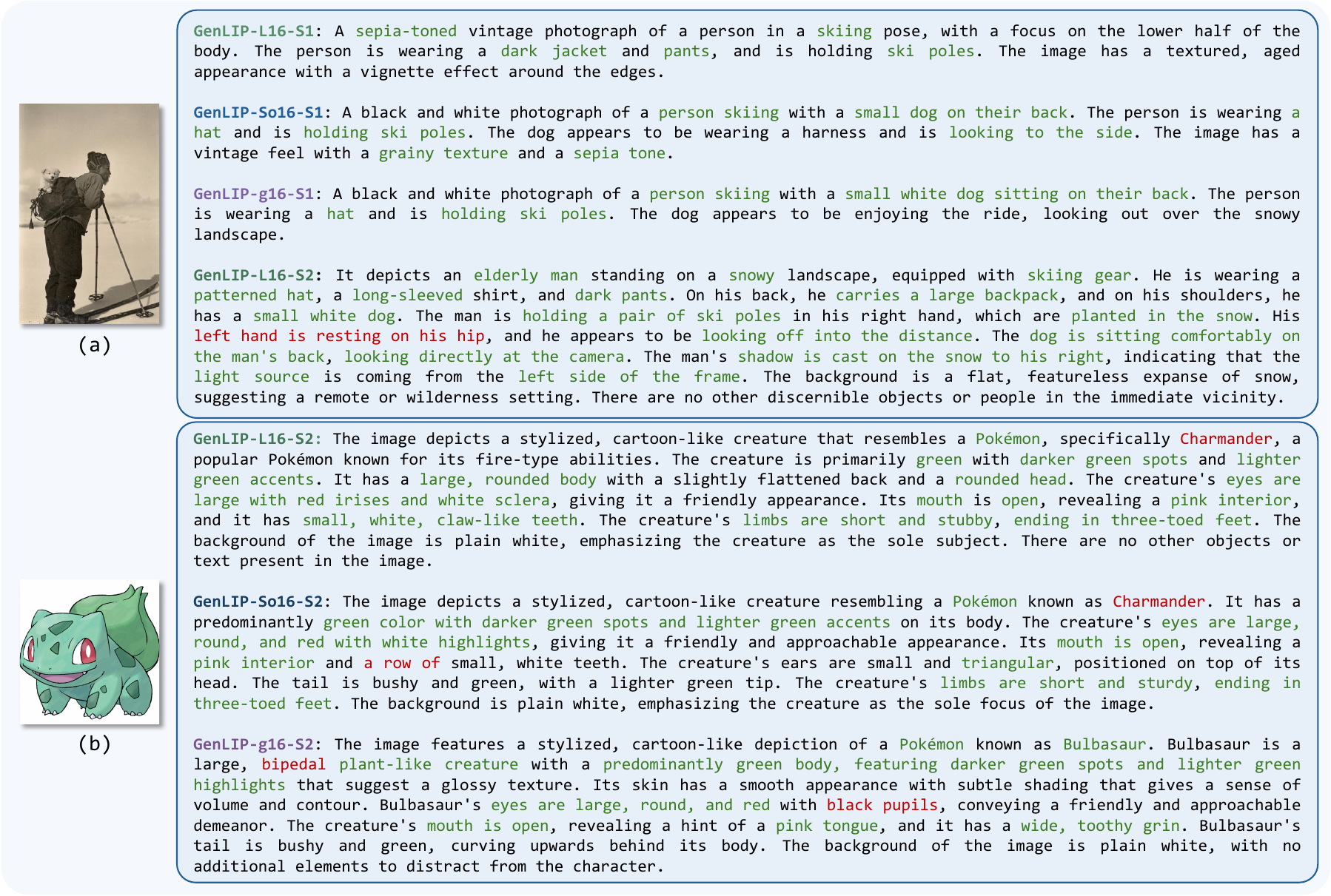}
  \caption{\textbf{Let ViT Speak.} We prompt GenLIP with ``Describe the image in details.'' and show representative generations. The first case compares three stage-1 models (\textcolor{l16s1}{GenLIP-L16-S1}, \textcolor{so16s1}{GenLIP-So16-S1}, and \textcolor{g16s1}{GenLIP-g16-S1}) with one stage-2 model (\textcolor{l16s2}{GenLIP-L16-S2}); the second case shows three stage-2 models. Green and red text indicate correct and incorrect key content, respectively.}
  \label{fig:caption-1}
\end{figure}

As shown in Figure~\ref{fig:caption-1}, GenLIP already produces fluent and semantically grounded descriptions. From stage 1 to stage 2, the responses become longer and more detailed, which is consistent with the finer-grained caption data used in the second pretraining stage. The captioning ability also improves with model scale. In the second example, the two smaller models, GenLIP-L16 and GenLIP-So16, mistake ``Bulbasaur'' for ``Charmander'', whereas the largest model, GenLIP-g16, identifies it correctly and provides richer details.

\begin{figure}[!htb]
  \centering
  \includegraphics[width=\linewidth]{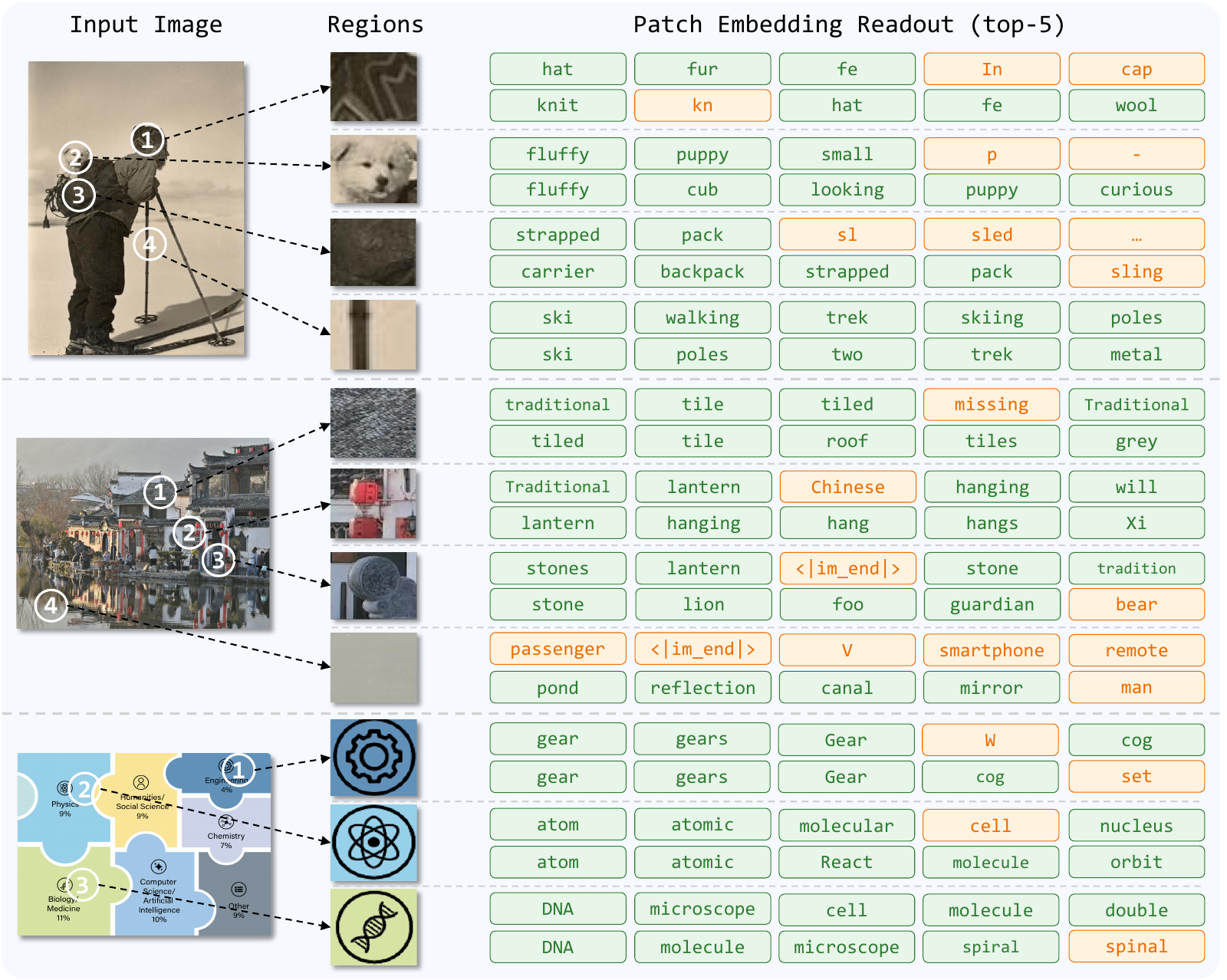}
  \caption{\textbf{Patch Semantics Readout.} We directly unembed selected image patch features with the language modeling head to inspect the language concepts aligned with local regions. For each case, we show 3--4 regions for \textcolor{g16s1}{GenLIP-g16-S1} (top row) and \textcolor{g16s2}{GenLIP-g16-S2} (bottom row), together with the top-5 predicted tokens from left to right. Green boxes indicate related tokens and yellow boxes indicate unrelated ones.}
  \label{fig:patchsemantics-1}
\end{figure}

\subsubsection{Patch Semantics Readout}
Beyond direct caption generation, we also probe what individual image patch features represent by translating them into language tokens with model's language modeling head. As shown in Figure~\ref{fig:patchsemantics-1}, GenLIP spontaneously aligns some local visual regions with meaningful language concepts, an emergent property learned during pretraining.
In the examples shown, both \textcolor{g16s1}{GenLIP-g16-S1} and \textcolor{g16s2}{GenLIP-g16-S2} models associate selected regions with semantically relevant concepts ranging from natural objects to abstract patterns. The \textcolor{g16s2}{GenLIP-g16-S2} model exhibits stronger alignment in both semantic correctness and relevance, likely due to the finer-grained captions and higher-quality images used in the second pretraining stage. Interestingly, this behavior is only observed in the two larger models, GenLIP-So16 and GenLIP-g16, with the latter showing more stable alignment. After stage 2, the readout semantics generally becomes more closely matched to the selected image regions. Although no explicit visual supervision is used, the model still learns to associate image patches with corresponding language concepts through generative pretraining on image-caption data.

Overall, the caption generation and patch-semantics experiments show that GenLIP can jointly model and align visual and linguistic modality, supporting its use as a strong vision encoder for MLLMs.

Additional qualitative examples, evaluation details, and a detailed discussion of attention sink are provided in the appendix.

\section{Conclusions}
This work presents GenLIP, a simple generative vision-language pretraining method based on a unified transformer architecture and a standard language modeling objective.
By jointly modeling visual and textual inputs with a single transformer, GenLIP aligns the two modalities through early fusion and directly optimizes the vision backbone for generative language prediction.
Despite its architectural and objective simplicity, GenLIP demonstrates strong data efficiency and scalability for vision-language pretraining, achieving competitive or superior performance across a wide range of multimodal benchmarks with substantially less training data than strong baselines.
We hope our exploration of generative vision-language pretraining will inspire future research toward more effective and scalable multimodal learning.

\paragraph{\textbf{Limitations.}}
Several limitations warrant consideration:
(i) our validation experiments are conducted on an academic-scale MLLM setting, LLaVA-NeXT, and the generalizability to cutting-edge MLLMs remains to be verified;
(ii) the pretraining dataset is limited to $1.0$B scale, the scaling behavior at even larger volumes is yet to be explored;
(iii) the reliance on high-quality captions introduces significant data acquisition costs.

\section{Acknowledgments}
This work was mainly sponsored by the National Natural Science Foundation of China (No.92470203).

\clearpage

\bibliographystyle{plainnat}
\bibliography{main}

\begin{thebibliography}{94}
\providecommand{\natexlab}[1]{#1}
\providecommand{\url}[1]{\texttt{#1}}
\expandafter\ifx\csname urlstyle\endcsname\relax
  \providecommand{\doi}[1]{doi: #1}\else
  \providecommand{\doi}{doi: \begingroup \urlstyle{rm}\Url}\fi

\bibitem[Achiam et~al.(2023)Achiam, Adler, Agarwal, Ahmad, Akkaya, Aleman,
  Almeida, Altenschmidt, Altman, Anadkat, et~al.]{achiam2023gpt}
Josh Achiam, Steven Adler, Sandhini Agarwal, Lama Ahmad, Ilge Akkaya,
  Florencia~Leoni Aleman, Diogo Almeida, Janko Altenschmidt, Sam Altman,
  Shyamal Anadkat, et~al.
\newblock Gpt-4 technical report.
\newblock \emph{arXiv preprint arXiv:2303.08774}, 2023.

\bibitem[Agrawal et~al.(2019)Agrawal, Desai, Wang, Chen, Jain, Johnson, Batra,
  Parikh, Lee, and Anderson]{agrawal2019nocaps}
Harsh Agrawal, Karan Desai, Yufei Wang, Xinlei Chen, Rishabh Jain, Mark
  Johnson, Dhruv Batra, Devi Parikh, Stefan Lee, and Peter Anderson.
\newblock Nocaps: Novel object captioning at scale.
\newblock In \emph{Proceedings of the IEEE/CVF international conference on
  computer vision}, pages 8948--8957, 2019.

\bibitem[Alayrac et~al.(2022)Alayrac, Donahue, Luc, Miech, Barr, Hasson, Lenc,
  Mensch, Millican, Reynolds, et~al.]{alayrac2022flamingo}
Jean-Baptiste Alayrac, Jeff Donahue, Pauline Luc, Antoine Miech, Iain Barr,
  Yana Hasson, Karel Lenc, Arthur Mensch, Katherine Millican, Malcolm Reynolds,
  et~al.
\newblock Flamingo: a visual language model for few-shot learning.
\newblock \emph{Advances in neural information processing systems},
  35:\penalty0 23716--23736, 2022.

\bibitem[An et~al.(2024)An, Yang, Dai, Feng, and Deng]{anxiang_2024_mlcd}
Xiang An, Kaicheng Yang, Xiangzi Dai, Ziyong Feng, and Jiankang Deng.
\newblock Multi-label cluster discrimination for visual representation
  learning.
\newblock In \emph{ECCV}, 2024.

\bibitem[An et~al.(2025)An, Zhao, Yu, Tang, and Wang]{an2025systematic}
Yongqi An, Xu~Zhao, Tao Yu, Ming Tang, and Jinqiao Wang.
\newblock Systematic outliers in large language models.
\newblock \emph{arXiv preprint arXiv:2502.06415}, 2025.

\bibitem[Bai et~al.(2023{\natexlab{a}})Bai, Bai, Chu, Cui, Dang, Deng, Fan, Ge,
  Han, Huang, et~al.]{bai2023qwen}
Jinze Bai, Shuai Bai, Yunfei Chu, Zeyu Cui, Kai Dang, Xiaodong Deng, Yang Fan,
  Wenbin Ge, Yu~Han, Fei Huang, et~al.
\newblock Qwen technical report.
\newblock \emph{arXiv preprint arXiv:2309.16609}, 2023{\natexlab{a}}.

\bibitem[Bai et~al.(2023{\natexlab{b}})Bai, Bai, Yang, Wang, Tan, Wang, Lin,
  Zhou, and Zhou]{qwenvl}
Jinze Bai, Shuai Bai, Shusheng Yang, Shijie Wang, Sinan Tan, Peng Wang, Junyang
  Lin, Chang Zhou, and Jingren Zhou.
\newblock Qwen-vl: A versatile vision-language model for understanding,
  localization, text reading, and beyond.
\newblock \emph{arXiv preprint arXiv:2308.12966}, 2023{\natexlab{b}}.

\bibitem[Bai et~al.(2025)Bai, Cai, Chen, Chen, Chen, Cheng, Deng, Ding, Gao,
  Ge, et~al.]{bai2025qwen3vl}
Shuai Bai, Yuxuan Cai, Ruizhe Chen, Keqin Chen, Xionghui Chen, Zesen Cheng,
  Lianghao Deng, Wei Ding, Chang Gao, Chunjiang Ge, et~al.
\newblock Qwen3-vl technical report.
\newblock \emph{arXiv preprint arXiv:2511.21631}, 2025.

\bibitem[Bao et~al.(2022)Bao, Wang, Dong, and Wei]{bao2022vl}
Hangbo Bao, Wenhui Wang, Li~Dong, and Furu Wei.
\newblock Vl-beit: Generative vision-language pretraining.
\newblock \emph{arXiv preprint arXiv:2206.01127}, 2022.

\bibitem[Beyer et~al.(2024)Beyer, Steiner, Pinto, Kolesnikov, Wang, Salz,
  Neumann, Alabdulmohsin, Tschannen, Bugliarello, et~al.]{beyer2024paligemma}
Lucas Beyer, Andreas Steiner, Andr{\'e}~Susano Pinto, Alexander Kolesnikov,
  Xiao Wang, Daniel Salz, Maxim Neumann, Ibrahim Alabdulmohsin, Michael
  Tschannen, Emanuele Bugliarello, et~al.
\newblock Paligemma: A versatile 3b vlm for transfer.
\newblock \emph{arXiv preprint arXiv:2407.07726}, 2024.

\bibitem[Chen et~al.(2025)Chen, Xu, Pan, Hu, Qin, Goldstein, Huang, Zhou, Xie,
  Savarese, et~al.]{chen2025blip3}
Jiuhai Chen, Zhiyang Xu, Xichen Pan, Yushi Hu, Can Qin, Tom Goldstein, Lifu
  Huang, Tianyi Zhou, Saining Xie, Silvio Savarese, et~al.
\newblock Blip3-o: A family of fully open unified multimodal
  models-architecture, training and dataset.
\newblock \emph{arXiv preprint arXiv:2505.09568}, 2025.

\bibitem[Chen et~al.(2024{\natexlab{a}})Chen, Wang, Peng, and Ji]{chen2024solo}
Yangyi Chen, Xingyao Wang, Hao Peng, and Heng Ji.
\newblock A single transformer for scalable vision-language modeling.
\newblock \emph{Transactions on Machine Learning Research}, 2024{\natexlab{a}}.

\bibitem[Chen et~al.(2024{\natexlab{b}})Chen, Wu, Wang, Su, Chen, Xing, Zhong,
  Zhang, Zhu, Lu, et~al.]{chen2024internvl}
Zhe Chen, Jiannan Wu, Wenhai Wang, Weijie Su, Guo Chen, Sen Xing, Muyan Zhong,
  Qinglong Zhang, Xizhou Zhu, Lewei Lu, et~al.
\newblock Internvl: Scaling up vision foundation models and aligning for
  generic visual-linguistic tasks.
\newblock In \emph{Proceedings of the IEEE/CVF conference on computer vision
  and pattern recognition}, pages 24185--24198, 2024{\natexlab{b}}.

\bibitem[Cherti et~al.(2023{\natexlab{a}})Cherti, Beaumont, Wightman, Wortsman,
  Ilharco, Gordon, Schuhmann, Schmidt, and Jitsev]{cherti2023openclip}
Mehdi Cherti, Romain Beaumont, Ross Wightman, Mitchell Wortsman, Gabriel
  Ilharco, Cade Gordon, Christoph Schuhmann, Ludwig Schmidt, and Jenia Jitsev.
\newblock Reproducible scaling laws for contrastive language-image learning.
\newblock In \emph{Proceedings of the IEEE/CVF conference on computer vision
  and pattern recognition}, pages 2818--2829, 2023{\natexlab{a}}.

\bibitem[Cherti et~al.(2023{\natexlab{b}})Cherti, Beaumont, Wightman, Wortsman,
  Ilharco, Gordon, Schuhmann, Schmidt, and Jitsev]{cherti2023reproducible}
Mehdi Cherti, Romain Beaumont, Ross Wightman, Mitchell Wortsman, Gabriel
  Ilharco, Cade Gordon, Christoph Schuhmann, Ludwig Schmidt, and Jenia Jitsev.
\newblock Reproducible scaling laws for contrastive language-image learning.
\newblock In \emph{Proceedings of the IEEE/CVF conference on computer vision
  and pattern recognition}, pages 2818--2829, 2023{\natexlab{b}}.

\bibitem[Chuang et~al.(2025)Chuang, Li, Wang, Yeh, Lyu, Raghavendra, Glass,
  HUANG, Weston, Zettlemoyer, et~al.]{chuang2025metaclip2}
Yung-Sung Chuang, Yang Li, Dong Wang, Ching-Feng Yeh, Kehan Lyu, Ramya
  Raghavendra, James~R Glass, LIFEI HUANG, Jason~E Weston, Luke Zettlemoyer,
  et~al.
\newblock Meta clip 2: A worldwide scaling recipe.
\newblock In \emph{The Thirty-ninth Annual Conference on Neural Information
  Processing Systems}, 2025.

\bibitem[Darcet et~al.(2023)Darcet, Oquab, Mairal, and
  Bojanowski]{darcet2023registers}
Timoth{\'e}e Darcet, Maxime Oquab, Julien Mairal, and Piotr Bojanowski.
\newblock Vision transformers need registers.
\newblock \emph{arXiv preprint arXiv:2309.16588}, 2023.

\bibitem[Deng et~al.(2009)Deng, Dong, Socher, Li, Li, and
  Fei-Fei]{deng2009imagenet}
Jia Deng, Wei Dong, Richard Socher, Li-Jia Li, Kai Li, and Li~Fei-Fei.
\newblock Imagenet: A large-scale hierarchical image database.
\newblock In \emph{2009 IEEE conference on computer vision and pattern
  recognition}, pages 248--255. Ieee, 2009.

\bibitem[Diao et~al.(2024)Diao, Cui, Li, Wang, Lu, and Wang]{diao2024eve}
Haiwen Diao, Yufeng Cui, Xiaotong Li, Yueze Wang, Huchuan Lu, and Xinlong Wang.
\newblock Unveiling encoder-free vision-language models.
\newblock \emph{Advances in Neural Information Processing Systems},
  37:\penalty0 52545--52567, 2024.

\bibitem[Diao et~al.(2025{\natexlab{a}})Diao, Li, Wu, Dai, Wang, Deng, Lu, Lin,
  and Liu]{diao2025pixels}
Haiwen Diao, Mingxuan Li, Silei Wu, Linjun Dai, Xiaohua Wang, Hanming Deng,
  Lewei Lu, Dahua Lin, and Ziwei Liu.
\newblock From pixels to words--towards native vision-language primitives at
  scale.
\newblock \emph{arXiv preprint arXiv:2510.14979}, 2025{\natexlab{a}}.

\bibitem[Diao et~al.(2025{\natexlab{b}})Diao, Li, Cui, Wang, Deng, Pan, Wang,
  Lu, and Wang]{diao2025evev2}
Haiwen Diao, Xiaotong Li, Yufeng Cui, Yueze Wang, Haoge Deng, Ting Pan, Wenxuan
  Wang, Huchuan Lu, and Xinlong Wang.
\newblock Evev2: Improved baselines for encoder-free vision-language models.
\newblock In \emph{Proceedings of the IEEE/CVF International Conference on
  Computer Vision}, pages 21014--21025, 2025{\natexlab{b}}.

\bibitem[Dosovitskiy et~al.(2021)Dosovitskiy, Beyer, Kolesnikov, Weissenborn,
  Zhai, Unterthiner, Dehghani, Minderer, Heigold, Gelly, Uszkoreit, and
  Houlsby]{alexander2021vit}
Alexey Dosovitskiy, Lucas Beyer, Alexander Kolesnikov, Dirk Weissenborn,
  Xiaohua Zhai, Thomas Unterthiner, Mostafa Dehghani, Matthias Minderer, Georg
  Heigold, Sylvain Gelly, Jakob Uszkoreit, and Neil Houlsby.
\newblock An image is worth 16x16 words: Transformers for image recognition at
  scale.
\newblock In \emph{9th International Conference on Learning Representations,
  {ICLR} 2021}. OpenReview.net, 2021.

\bibitem[Fan et~al.(2023)Fan, Krishnan, Isola, Katabi, and
  Tian]{fan2023improving}
Lijie Fan, Dilip Krishnan, Phillip Isola, Dina Katabi, and Yonglong Tian.
\newblock Improving clip training with language rewrites.
\newblock \emph{Advances in Neural Information Processing Systems},
  36:\penalty0 35544--35575, 2023.

\bibitem[Fini et~al.(2025)Fini, Shukor, Li, Dufter, Klein, Haldimann,
  Aitharaju, da~Costa, B{\'e}thune, Gan, et~al.]{fini2025aimv2}
Enrico Fini, Mustafa Shukor, Xiujun Li, Philipp Dufter, Michal Klein, David
  Haldimann, Sai Aitharaju, Victor G~Turrisi da~Costa, Louis B{\'e}thune, Zhe
  Gan, et~al.
\newblock Multimodal autoregressive pre-training of large vision encoders.
\newblock In \emph{Proceedings of the Computer Vision and Pattern Recognition
  Conference}, pages 9641--9654, 2025.

\bibitem[Fu et~al.(2023)Fu, Chen, Shen, Qin, Zhang, Lin, Yang, Zheng, Li, Sun,
  et~al.]{fu2023mme}
Chaoyou Fu, Peixian Chen, Yunhang Shen, Yulei Qin, Mengdan Zhang, Xu~Lin,
  Jinrui Yang, Xiawu Zheng, Ke~Li, Xing Sun, et~al.
\newblock Mme: A comprehensive evaluation benchmark for multimodal large
  language models.
\newblock \emph{arXiv preprint arXiv:2306.13394}, 2023.

\bibitem[Gadre et~al.(2023)Gadre, Ilharco, Fang, Hayase, Smyrnis, Nguyen,
  Marten, Wortsman, Ghosh, Zhang, et~al.]{gadre2023datacomp}
Samir~Yitzhak Gadre, Gabriel Ilharco, Alex Fang, Jonathan Hayase, Georgios
  Smyrnis, Thao Nguyen, Ryan Marten, Mitchell Wortsman, Dhruba Ghosh, Jieyu
  Zhang, et~al.
\newblock Datacomp: In search of the next generation of multimodal datasets.
\newblock \emph{Advances in Neural Information Processing Systems},
  36:\penalty0 27092--27112, 2023.

\bibitem[Goyal et~al.(2017)Goyal, Khot, Summers-Stay, Batra, and
  Parikh]{goyal2017vqav2}
Yash Goyal, Tejas Khot, Douglas Summers-Stay, Dhruv Batra, and Devi Parikh.
\newblock Making the v in vqa matter: Elevating the role of image understanding
  in visual question answering.
\newblock In \emph{Proceedings of the IEEE conference on computer vision and
  pattern recognition}, pages 6904--6913, 2017.

\bibitem[Gu et~al.(2024)Gu, Zhang, Zhou, Yu, Xing, Wang, Cao, Jia, Zhang, Wang,
  et~al.]{gu2024infinity}
Shuhao Gu, Jialing Zhang, Siyuan Zhou, Kevin Yu, Zhaohu Xing, Liangdong Wang,
  Zhou Cao, Jintao Jia, Zhuoyi Zhang, Yixuan Wang, et~al.
\newblock Infinity-mm: Scaling multimodal performance with large-scale and
  high-quality instruction data.
\newblock \emph{CoRR}, 2024.

\bibitem[Huang et~al.(2024)Huang, Ye, Kang, Feng, and
  Fan]{huang2024classification}
Zilong Huang, Qinghao Ye, Bingyi Kang, Jiashi Feng, and Haoqi Fan.
\newblock Classification done right for vision-language pre-training.
\newblock \emph{Advances in Neural Information Processing Systems},
  37:\penalty0 96483--96504, 2024.

\bibitem[Hudson and Manning(2019)]{hudson2019gqa}
Drew~A Hudson and Christopher~D Manning.
\newblock Gqa: A new dataset for real-world visual reasoning and compositional
  question answering.
\newblock In \emph{Proceedings of the IEEE/CVF conference on computer vision
  and pattern recognition}, pages 6700--6709, 2019.

\bibitem[Jang et~al.(2023)Jang, Kong, Jeon, Kim, and Kwak]{jang2023unifying}
Jiho Jang, Chaerin Kong, Donghyeon Jeon, Seonhoon Kim, and Nojun Kwak.
\newblock Unifying vision-language representation space with single-tower
  transformer.
\newblock In \emph{Proceedings of the AAAI conference on artificial
  intelligence}, volume~37, pages 980--988, 2023.

\bibitem[Jia et~al.(2021)Jia, Yang, Xia, Chen, Parekh, Pham, Le, Sung, Li, and
  Duerig]{jia2021scaling}
Chao Jia, Yinfei Yang, Ye~Xia, Yi-Ting Chen, Zarana Parekh, Hieu Pham, Quoc Le,
  Yun-Hsuan Sung, Zhen Li, and Tom Duerig.
\newblock Scaling up visual and vision-language representation learning with
  noisy text supervision.
\newblock In \emph{International conference on machine learning}, pages
  4904--4916. PMLR, 2021.

\bibitem[Kembhavi et~al.(2016)Kembhavi, Salvato, Kolve, Seo, Hajishirzi, and
  Farhadi]{kembhavi2016ai2d}
Aniruddha Kembhavi, Mike Salvato, Eric Kolve, Minjoon Seo, Hannaneh Hajishirzi,
  and Ali Farhadi.
\newblock A diagram is worth a dozen images.
\newblock In \emph{European conference on computer vision}, pages 235--251.
  Springer, 2016.

\bibitem[Lai et~al.(2024)Lai, Zhang, Zhang, Wu, Bai, Timofeev, Du, Gan, Shan,
  Chuah, et~al.]{lai2024veclip}
Zhengfeng Lai, Haotian Zhang, Bowen Zhang, Wentao Wu, Haoping Bai, Aleksei
  Timofeev, Xianzhi Du, Zhe Gan, Jiulong Shan, Chen-Nee Chuah, et~al.
\newblock Veclip: Improving clip training via visual-enriched captions.
\newblock In \emph{European Conference on Computer Vision}, pages 111--127.
  Springer, 2024.

\bibitem[Lei et~al.(2025)Lei, Wang, Wang, Li, Liew, Feng, and
  Huang]{lei2025sail}
Weixian Lei, Jiacong Wang, Haochen Wang, Xiangtai Li, Jun~Hao Liew, Jiashi
  Feng, and Zilong Huang.
\newblock The scalability of simplicity: Empirical analysis of vision-language
  learning with a single transformer.
\newblock In \emph{Proceedings of the IEEE/CVF International Conference on
  Computer Vision (ICCV)}, pages 20758--20769, October 2025.

\bibitem[Li et~al.(2024{\natexlab{a}})Li, Zhang, Guo, Zhang, Li, Zhang, Zhang,
  Li, Liu, and Li]{li2024llavaonevision}
Bo~Li, Yuanhan Zhang, Dong Guo, Renrui Zhang, Feng Li, Hao Zhang, Kaichen
  Zhang, Yanwei Li, Ziwei Liu, and Chunyuan Li.
\newblock Llava-onevision: Easy visual task transfer.
\newblock \emph{CoRR}, 2024{\natexlab{a}}.

\bibitem[Li et~al.(2024{\natexlab{b}})Li, Ge, Chen, Ge, Zhang, and
  Shan]{li2024seedbench2}
Bohao Li, Yuying Ge, Yi~Chen, Yixiao Ge, Ruimao Zhang, and Ying Shan.
\newblock Seed-bench-2-plus: Benchmarking multimodal large language models with
  text-rich visual comprehension.
\newblock \emph{CoRR}, 2024{\natexlab{b}}.

\bibitem[Li et~al.(2024{\natexlab{c}})Li, Ge, Ge, Wang, Wang, Zhang, and
  Shan]{li2024seednbench}
Bohao Li, Yuying Ge, Yixiao Ge, Guangzhi Wang, Rui Wang, Ruimao Zhang, and Ying
  Shan.
\newblock Seed-bench: Benchmarking multimodal large language models.
\newblock In \emph{Proceedings of the IEEE/CVF Conference on Computer Vision
  and Pattern Recognition}, pages 13299--13308, 2024{\natexlab{c}}.

\bibitem[Li et~al.(2021)Li, Selvaraju, Gotmare, Joty, Xiong, and
  Hoi]{li2021align}
Junnan Li, Ramprasaath Selvaraju, Akhilesh Gotmare, Shafiq Joty, Caiming Xiong,
  and Steven Chu~Hong Hoi.
\newblock Align before fuse: Vision and language representation learning with
  momentum distillation.
\newblock \emph{Advances in neural information processing systems},
  34:\penalty0 9694--9705, 2021.

\bibitem[Li et~al.(2022)Li, Li, Xiong, and Hoi]{li2022blip}
Junnan Li, Dongxu Li, Caiming Xiong, and Steven Hoi.
\newblock Blip: Bootstrapping language-image pre-training for unified
  vision-language understanding and generation.
\newblock In \emph{International conference on machine learning}, pages
  12888--12900. PMLR, 2022.

\bibitem[Li et~al.(2025{\natexlab{a}})Li, Zhang, Li, Yuan, Chen, Zhou, Meng,
  Sun, Xu, Qi, et~al.]{li2025denseworld}
Xiangtai Li, Tao Zhang, Yanwei Li, Haobo Yuan, Shihao Chen, Yikang Zhou, Jiahao
  Meng, Yueyi Sun, Shilin Xu, Lu~Qi, et~al.
\newblock Denseworld-1m: Towards detailed dense grounded caption in the real
  world.
\newblock \emph{arXiv preprint arXiv:2506.24102}, 2025{\natexlab{a}}.

\bibitem[Li et~al.(2024{\natexlab{d}})Li, Tu, Hui, Wang, Zhao, Xiao, Ren, Mei,
  Liu, Zheng, et~al.]{xianhang2024recap}
Xianhang Li, Haoqin Tu, Mude Hui, Zeyu Wang, Bingchen Zhao, Junfei Xiao,
  Sucheng Ren, Jieru Mei, Qing Liu, Huangjie Zheng, et~al.
\newblock What if we recaption billions of web images with llama-3?
\newblock In \emph{International Conference on Machine Learning}. PMLR,
  2024{\natexlab{d}}.

\bibitem[Li et~al.(2025{\natexlab{b}})Li, Liu, Tu, and Xie]{li2025openvision}
Xianhang Li, Yanqing Liu, Haoqin Tu, and Cihang Xie.
\newblock Openvision: A fully-open, cost-effective family of advanced vision
  encoders for multimodal learning.
\newblock In \emph{Proceedings of the IEEE/CVF International Conference on
  Computer Vision}, pages 3977--3987, 2025{\natexlab{b}}.

\bibitem[Li et~al.(2024{\natexlab{e}})Li, Zhang, Diao, Wang, Wang, and
  Duan]{li2024densefusion}
Xiaotong Li, Fan Zhang, Haiwen Diao, Yueze Wang, Xinlong Wang, and Lingyu Duan.
\newblock Densefusion-1m: Merging vision experts for comprehensive multimodal
  perception.
\newblock \emph{Advances in Neural Information Processing Systems},
  37:\penalty0 18535--18556, 2024{\natexlab{e}}.

\bibitem[Lin et~al.(2024)Lin, Chen, Pathak, Zhang, and Ramanan]{linrevisiting}
Zhiqiu Lin, Xinyue Chen, Deepak Pathak, Pengchuan Zhang, and Deva Ramanan.
\newblock Revisiting the role of language priors in vision-language models.
\newblock In \emph{Forty-first International Conference on Machine Learning},
  2024.

\bibitem[Liu et~al.(2023)Liu, Li, Wu, and Lee]{liu2023llava}
Haotian Liu, Chunyuan Li, Qingyang Wu, and Yong~Jae Lee.
\newblock Visual instruction tuning.
\newblock \emph{Advances in neural information processing systems},
  36:\penalty0 34892--34916, 2023.

\bibitem[Liu et~al.(2024{\natexlab{a}})Liu, Li, Li, Li, Zhang, Shen, and
  Lee]{liu2024llavanext}
Haotian Liu, Chunyuan Li, Yuheng Li, Bo~Li, Yuanhan Zhang, Sheng Shen, and
  Yong~Jae Lee.
\newblock Llava-next: Improved reasoning, ocr, and world knowledge, january
  2024.
\newblock \emph{URL https://llava-vl. github. io/blog/2024-01-30-llava-next},
  1\penalty0 (8), 2024{\natexlab{a}}.

\bibitem[Liu et~al.(2024{\natexlab{b}})Liu, Li, Wang, Zhao, and
  Xie]{liu2024clips}
Yanqing Liu, Xianhang Li, Zeyu Wang, Bingchen Zhao, and Cihang Xie.
\newblock Clips: An enhanced clip framework for learning with synthetic
  captions.
\newblock \emph{arXiv preprint arXiv:2411.16828}, 2024{\natexlab{b}}.

\bibitem[Liu et~al.(2025)Liu, Li, Zhang, Wang, Zheng, Zhou, and
  Xie]{liu2025openvision2}
Yanqing Liu, Xianhang Li, Letian Zhang, Zirui Wang, Zeyu Zheng, Yuyin Zhou, and
  Cihang Xie.
\newblock Openvision 2: A family of generative pretrained visual encoders for
  multimodal learning.
\newblock \emph{arXiv preprint arXiv:2509.01644}, 2025.

\bibitem[Liu et~al.(2024{\natexlab{c}})Liu, Duan, Zhang, Li, Zhang, Zhao, Yuan,
  Wang, He, Liu, et~al.]{liu2024mmbench}
Yuan Liu, Haodong Duan, Yuanhan Zhang, Bo~Li, Songyang Zhang, Wangbo Zhao, Yike
  Yuan, Jiaqi Wang, Conghui He, Ziwei Liu, et~al.
\newblock Mmbench: Is your multi-modal model an all-around player?
\newblock In \emph{European conference on computer vision}, pages 216--233.
  Springer, 2024{\natexlab{c}}.

\bibitem[Liu et~al.(2024{\natexlab{d}})Liu, Li, Huang, Yang, Yu, Li, Yin, Liu,
  Jin, and Bai]{liu2024ocrbench}
Yuliang Liu, Zhang Li, Mingxin Huang, Biao Yang, Wenwen Yu, Chunyuan Li,
  Xu-Cheng Yin, Cheng-Lin Liu, Lianwen Jin, and Xiang Bai.
\newblock Ocrbench: on the hidden mystery of ocr in large multimodal models.
\newblock \emph{Science China Information Sciences}, 67\penalty0 (12):\penalty0
  220102, 2024{\natexlab{d}}.

\bibitem[Lu et~al.(2022)Lu, Mishra, Xia, Qiu, Chang, Zhu, Tafjord, Clark, and
  Kalyan]{lu2022scienceqa}
Pan Lu, Swaroop Mishra, Tony Xia, Liang Qiu, Kai-Wei Chang, Song-Chun Zhu,
  Oyvind Tafjord, Peter Clark, and Ashwin Kalyan.
\newblock Learn to explain: Multimodal reasoning via thought chains for science
  question answering.
\newblock In \emph{The 36th Conference on Neural Information Processing Systems
  (NeurIPS)}, 2022.

\bibitem[Lu et~al.(2024)Lu, Bansal, Xia, Liu, Li, Hajishirzi, Cheng, Chang,
  Galley, and Gao]{lu2024mathvista}
Pan Lu, Hritik Bansal, Tony Xia, Jiacheng Liu, Chunyuan Li, Hannaneh
  Hajishirzi, Hao Cheng, Kai-Wei Chang, Michel Galley, and Jianfeng Gao.
\newblock Mathvista: Evaluating mathematical reasoning of foundation models in
  visual contexts.
\newblock In \emph{International Conference on Learning Representations
  (ICLR)}, 2024.

\bibitem[Mao et~al.(2016)Mao, Huang, Toshev, Camburu, Yuille, and
  Murphy]{mao2016generation}
Junhua Mao, Jonathan Huang, Alexander Toshev, Oana Camburu, Alan~L Yuille, and
  Kevin Murphy.
\newblock Generation and comprehension of unambiguous object descriptions.
\newblock In \emph{Proceedings of the IEEE conference on computer vision and
  pattern recognition}, pages 11--20, 2016.

\bibitem[Masry et~al.(2022)Masry, Do, Tan, Joty, and Hoque]{masry2022chartqa}
Ahmed Masry, Xuan~Long Do, Jia~Qing Tan, Shafiq Joty, and Enamul Hoque.
\newblock Chartqa: A benchmark for question answering about charts with visual
  and logical reasoning.
\newblock In \emph{Findings of the association for computational linguistics:
  ACL 2022}, pages 2263--2279, 2022.

\bibitem[Mathew et~al.(2021)Mathew, Karatzas, and Jawahar]{mathew2021docvqa}
Minesh Mathew, Dimosthenis Karatzas, and CV~Jawahar.
\newblock Docvqa: A dataset for vqa on document images.
\newblock In \emph{Proceedings of the IEEE/CVF winter conference on
  applications of computer vision}, pages 2200--2209, 2021.

\bibitem[Mathew et~al.(2022)Mathew, Bagal, Tito, Karatzas, Valveny, and
  Jawahar]{mathew2022infographicvqa}
Minesh Mathew, Viraj Bagal, Rub{\`e}n Tito, Dimosthenis Karatzas, Ernest
  Valveny, and CV~Jawahar.
\newblock Infographicvqa.
\newblock In \emph{Proceedings of the IEEE/CVF Winter Conference on
  Applications of Computer Vision}, pages 1697--1706, 2022.

\bibitem[Oquab et~al.(2024)Oquab, Darcet, Moutakanni, Vo, Szafraniec, Khalidov,
  Fernandez, Haziza, Massa, El-Nouby, et~al.]{oquab2024dinov2}
Maxime Oquab, Timoth{\'e}e Darcet, Th{\'e}o Moutakanni, Huy Vo, Marc
  Szafraniec, Vasil Khalidov, Pierre Fernandez, Daniel Haziza, Francisco Massa,
  Alaaeldin El-Nouby, et~al.
\newblock Dinov2: Learning robust visual features without supervision.
\newblock \emph{Transactions on Machine Learning Research Journal}, 2024.

\bibitem[Qiu et~al.(2025)Qiu, Wang, Zheng, Huang, Wen, Yang, Men, Yu, Huang,
  Huang, et~al.]{qiu2025gated}
Zihan Qiu, Zekun Wang, Bo~Zheng, Zeyu Huang, Kaiyue Wen, Songlin Yang, Rui Men,
  Le~Yu, Fei Huang, Suozhi Huang, et~al.
\newblock Gated attention for large language models: Non-linearity, sparsity,
  and attention-sink-free.
\newblock \emph{arXiv preprint arXiv:2505.06708}, 2025.

\bibitem[Qwen et~al.(2025)Qwen, :, Yang, Yang, Zhang, Hui, Zheng, Yu, Li, Liu,
  Huang, Wei, Lin, Yang, Tu, Zhang, Yang, Yang, Zhou, Lin, Dang, Lu, Bao, Yang,
  Yu, Li, Xue, Zhang, Zhu, Men, Lin, Li, Tang, Xia, Ren, Ren, Fan, Su, Zhang,
  Wan, Liu, Cui, Zhang, and Qiu]{qwen2025qwen25technicalreport}
Qwen, :, An~Yang, Baosong Yang, Beichen Zhang, Binyuan Hui, Bo~Zheng, Bowen Yu,
  Chengyuan Li, Dayiheng Liu, Fei Huang, Haoran Wei, Huan Lin, Jian Yang,
  Jianhong Tu, Jianwei Zhang, Jianxin Yang, Jiaxi Yang, Jingren Zhou, Junyang
  Lin, Kai Dang, Keming Lu, Keqin Bao, Kexin Yang, Le~Yu, Mei Li, Mingfeng Xue,
  Pei Zhang, Qin Zhu, Rui Men, Runji Lin, Tianhao Li, Tianyi Tang, Tingyu Xia,
  Xingzhang Ren, Xuancheng Ren, Yang Fan, Yang Su, Yichang Zhang, Yu~Wan,
  Yuqiong Liu, Zeyu Cui, Zhenru Zhang, and Zihan Qiu.
\newblock Qwen2.5 technical report, 2025.
\newblock URL \url{https://arxiv.org/abs/2412.15115}.

\bibitem[Radford et~al.(2021)Radford, Kim, Hallacy, Ramesh, Goh, Agarwal,
  Sastry, Askell, Mishkin, Clark, et~al.]{radford2021clip}
Alec Radford, Jong~Wook Kim, Chris Hallacy, Aditya Ramesh, Gabriel Goh,
  Sandhini Agarwal, Girish Sastry, Amanda Askell, Pamela Mishkin, Jack Clark,
  et~al.
\newblock Learning transferable visual models from natural language
  supervision.
\newblock In \emph{International conference on machine learning}, pages
  8748--8763. PMLR, 2021.

\bibitem[Raffel et~al.(2020)Raffel, Shazeer, Roberts, Lee, Narang, Matena,
  Zhou, Li, and Liu]{raffel2020exploring}
Colin Raffel, Noam Shazeer, Adam Roberts, Katherine Lee, Sharan Narang, Michael
  Matena, Yanqi Zhou, Wei Li, and Peter~J Liu.
\newblock Exploring the limits of transfer learning with a unified text-to-text
  transformer.
\newblock \emph{Journal of machine learning research}, 21\penalty0
  (140):\penalty0 1--67, 2020.

\bibitem[Sidorov et~al.(2020)Sidorov, Hu, Rohrbach, and
  Singh]{sidorov2020textcaps}
Oleksii Sidorov, Ronghang Hu, Marcus Rohrbach, and Amanpreet Singh.
\newblock Textcaps: a dataset for image captioning with reading comprehension.
\newblock In \emph{European conference on computer vision}, pages 742--758.
  Springer, 2020.

\bibitem[Sim{\'e}oni et~al.(2025)Sim{\'e}oni, Vo, Seitzer, Baldassarre, Oquab,
  Jose, Khalidov, Szafraniec, Yi, Ramamonjisoa, Massa, Haziza, Wehrstedt, Wang,
  Darcet, Moutakanni, Sentana, Roberts, Vedaldi, Tolan, Brandt, Couprie,
  Mairal, J{\'e}gou, Labatut, and Bojanowski]{simeoni2025dinov3}
Oriane Sim{\'e}oni, Huy~V. Vo, Maximilian Seitzer, Federico Baldassarre, Maxime
  Oquab, Cijo Jose, Vasil Khalidov, Marc Szafraniec, Seungeun Yi, Micha{\"e}l
  Ramamonjisoa, Francisco Massa, Daniel Haziza, Luca Wehrstedt, Jianyuan Wang,
  Timoth{\'e}e Darcet, Th{\'e}o Moutakanni, Leonel Sentana, Claire Roberts,
  Andrea Vedaldi, Jamie Tolan, John Brandt, Camille Couprie, Julien Mairal,
  Herv{\'e} J{\'e}gou, Patrick Labatut, and Piotr Bojanowski.
\newblock {DINOv3}, 2025.
\newblock URL \url{https://arxiv.org/abs/2508.10104}.

\bibitem[Singh et~al.(2019)Singh, Natarajan, Shah, Jiang, Chen, Batra, Parikh,
  and Rohrbach]{singh2019textvqa}
Amanpreet Singh, Vivek Natarajan, Meet Shah, Yu~Jiang, Xinlei Chen, Dhruv
  Batra, Devi Parikh, and Marcus Rohrbach.
\newblock Towards vqa models that can read.
\newblock In \emph{Proceedings of the IEEE/CVF conference on computer vision
  and pattern recognition}, pages 8317--8326, 2019.

\bibitem[Sun et~al.(2024)Sun, Cui, Zhang, Zhang, Yu, Wang, Rao, Liu, Huang, and
  Wang]{sun2024generative}
Quan Sun, Yufeng Cui, Xiaosong Zhang, Fan Zhang, Qiying Yu, Yueze Wang,
  Yongming Rao, Jingjing Liu, Tiejun Huang, and Xinlong Wang.
\newblock Generative multimodal models are in-context learners.
\newblock In \emph{Proceedings of the IEEE/CVF Conference on Computer Vision
  and Pattern Recognition}, pages 14398--14409, 2024.

\bibitem[Team(2024{\natexlab{a}})]{team2024chameleon}
Chameleon Team.
\newblock Chameleon: Mixed-modal early-fusion foundation models.
\newblock \emph{arXiv preprint arXiv:2405.09818}, 2024{\natexlab{a}}.

\bibitem[Team et~al.(2025)Team, Du, Yin, Xing, Qu, Wang, Chen, Zhang, Du, Wei,
  et~al.]{team2025kimivl}
Kimi Team, Angang Du, Bohong Yin, Bowei Xing, Bowen Qu, Bowen Wang, Cheng Chen,
  Chenlin Zhang, Chenzhuang Du, Chu Wei, et~al.
\newblock Kimi-vl technical report.
\newblock \emph{arXiv preprint arXiv:2504.07491}, 2025.

\bibitem[Team et~al.(2026)Team, Bai, Bai, Bao, Cai, Cao, Charles, Che, Chen,
  Chen, et~al.]{team2026kimi-k2.5}
Kimi Team, Tongtong Bai, Yifan Bai, Yiping Bao, SH~Cai, Yuan Cao, Y~Charles,
  HS~Che, Cheng Chen, Guanduo Chen, et~al.
\newblock Kimi k2. 5: Visual agentic intelligence.
\newblock \emph{arXiv preprint arXiv:2602.02276}, 2026.

\bibitem[Team(2024{\natexlab{b}})]{qwen2.5}
Qwen Team.
\newblock Qwen2.5: A party of foundation models, September 2024{\natexlab{b}}.
\newblock URL \url{https://qwenlm.github.io/blog/qwen2.5/}.

\bibitem[Tong et~al.(2024{\natexlab{a}})Tong, Brown, Wu, Woo, Iyer, Akula,
  Yang, Yang, Middepogu, Wang, et~al.]{tong2024cambrian}
Peter Tong, Ellis Brown, Penghao Wu, Sanghyun Woo, Adithya Jairam~Vedagiri
  Iyer, Sai~Charitha Akula, Shusheng Yang, Jihan Yang, Manoj Middepogu, Ziteng
  Wang, et~al.
\newblock Cambrian-1: A fully open, vision-centric exploration of multimodal
  llms.
\newblock \emph{Advances in Neural Information Processing Systems},
  37:\penalty0 87310--87356, 2024{\natexlab{a}}.

\bibitem[Tong et~al.(2024{\natexlab{b}})Tong, Liu, Zhai, Ma, LeCun, and
  Xie]{tong2024eyes}
Shengbang Tong, Zhuang Liu, Yuexiang Zhai, Yi~Ma, Yann LeCun, and Saining Xie.
\newblock Eyes wide shut? exploring the visual shortcomings of multimodal llms,
  2024{\natexlab{b}}.

\bibitem[Touvron et~al.(2023)Touvron, Lavril, Izacard, Martinet, Lachaux,
  Lacroix, Rozi{\`e}re, Goyal, Hambro, Azhar, et~al.]{touvron2023llama}
Hugo Touvron, Thibaut Lavril, Gautier Izacard, Xavier Martinet, Marie-Anne
  Lachaux, Timoth{\'e}e Lacroix, Baptiste Rozi{\`e}re, Naman Goyal, Eric
  Hambro, Faisal Azhar, et~al.
\newblock Llama: Open and efficient foundation language models.
\newblock \emph{arXiv preprint arXiv:2302.13971}, 2023.

\bibitem[Tschannen et~al.(2023)Tschannen, Kumar, Steiner, Zhai, Houlsby, and
  Beyer]{tschannen2023cappa}
Michael Tschannen, Manoj Kumar, Andreas Steiner, Xiaohua Zhai, Neil Houlsby,
  and Lucas Beyer.
\newblock Image captioners are scalable vision learners too.
\newblock \emph{Advances in Neural Information Processing Systems},
  36:\penalty0 46830--46855, 2023.

\bibitem[Tschannen et~al.(2025)Tschannen, Gritsenko, Wang, Naeem,
  Alabdulmohsin, Parthasarathy, Evans, Beyer, Xia, Mustafa,
  et~al.]{tschannen2025siglip2}
Michael Tschannen, Alexey Gritsenko, Xiao Wang, Muhammad~Ferjad Naeem, Ibrahim
  Alabdulmohsin, Nikhil Parthasarathy, Talfan Evans, Lucas Beyer, Ye~Xia, Basil
  Mustafa, et~al.
\newblock Siglip 2: Multilingual vision-language encoders with improved
  semantic understanding, localization, and dense features.
\newblock \emph{arXiv preprint arXiv:2502.14786}, 2025.

\bibitem[Wang et~al.(2022)Wang, Yang, Hu, Li, Lin, Gan, Liu, Liu, and
  Wang]{wang2022git}
Jianfeng Wang, Zhengyuan Yang, Xiaowei Hu, Linjie Li, Kevin Lin, Zhe Gan,
  Zicheng Liu, Ce~Liu, and Lijuan Wang.
\newblock Git: A generative image-to-text transformer for vision and language.
\newblock \emph{arXiv preprint arXiv:2205.14100}, 2022.

\bibitem[Wang et~al.(2024)Wang, Bai, Tan, Wang, Fan, Bai, Chen, Liu, Wang, Ge,
  et~al.]{wang2024qwen2vl}
Peng Wang, Shuai Bai, Sinan Tan, Shijie Wang, Zhihao Fan, Jinze Bai, Keqin
  Chen, Xuejing Liu, Jialin Wang, Wenbin Ge, et~al.
\newblock Qwen2-vl: Enhancing vision-language model's perception of the world
  at any resolution.
\newblock \emph{arXiv preprint arXiv:2409.12191}, 2024.

\bibitem[Wang et~al.(2021)Wang, Yu, Yu, Dai, Tsvetkov, and Cao]{wang2021simvlm}
Zirui Wang, Jiahui Yu, Adams~Wei Yu, Zihang Dai, Yulia Tsvetkov, and Yuan Cao.
\newblock Simvlm: Simple visual language model pretraining with weak
  supervision.
\newblock \emph{arXiv preprint arXiv:2108.10904}, 2021.

\bibitem[{xAI}(2024)]{realworldqa2024}
{xAI}.
\newblock Realworldqa: A benchmark for real-world spatial understanding.
\newblock \url{https://huggingface.co/datasets/xai-org/RealworldQA}, 2024.
\newblock Accessed: 2025-04-26.

\bibitem[Xiao et~al.(2023)Xiao, Tian, Chen, Han, and Lewis]{xiao2023efficient}
Guangxuan Xiao, Yuandong Tian, Beidi Chen, Song Han, and Mike Lewis.
\newblock Efficient streaming language models with attention sinks.
\newblock \emph{arXiv preprint arXiv:2309.17453}, 2023.

\bibitem[Xie et~al.(2025)Xie, Yang, An, Wu, Zhao, Deng, Ran, Wang, Feng, Roy,
  Ismail, and Deng]{yinxie_2025_rice}
Yin Xie, Kaicheng Yang, Xiang An, Kun Wu, Yongle Zhao, Weimo Deng, Zimin Ran,
  Yumeng Wang, Ziyong Feng, Miles Roy, Elezi Ismail, and Jiankang Deng.
\newblock Region-based cluster discrimination for visual representation
  learning.
\newblock In \emph{ICCV}, 2025.

\bibitem[Xing et~al.(2025)Xing, Dong, Zang, Cao, Liang, Huang, Wang, Wu, and
  Lin]{xing2025caprl}
Long Xing, Xiaoyi Dong, Yuhang Zang, Yuhang Cao, Jianze Liang, Qidong Huang,
  Jiaqi Wang, Feng Wu, and Dahua Lin.
\newblock Caprl: Stimulating dense image caption capabilities via reinforcement
  learning.
\newblock \emph{arXiv preprint arXiv:2509.22647}, 2025.

\bibitem[Xu et~al.(2023)Xu, Xie, Tan, Huang, Howes, Sharma, Li, Ghosh,
  Zettlemoyer, and Feichtenhofer]{xu2023demystifying}
Hu~Xu, Saining Xie, Xiaoqing~Ellen Tan, Po-Yao Huang, Russell Howes, Vasu
  Sharma, Shang-Wen Li, Gargi Ghosh, Luke Zettlemoyer, and Christoph
  Feichtenhofer.
\newblock Demystifying clip data.
\newblock \emph{arXiv preprint arXiv:2309.16671}, 2023.

\bibitem[Yang et~al.(2025)Yang, Li, Yang, Zhang, Hui, Zheng, Yu, Gao, Huang,
  Lv, et~al.]{yang2025qwen3}
An~Yang, Anfeng Li, Baosong Yang, Beichen Zhang, Binyuan Hui, Bo~Zheng, Bowen
  Yu, Chang Gao, Chengen Huang, Chenxu Lv, et~al.
\newblock Qwen3 technical report.
\newblock \emph{arXiv preprint arXiv:2505.09388}, 2025.

\bibitem[Yang et~al.(2023)Yang, Deng, An, Li, Feng, Guo, Yang, and
  Liu]{yang2023alip}
Kaicheng Yang, Jiankang Deng, Xiang An, Jiawei Li, Ziyong Feng, Jia Guo, Jing
  Yang, and Tongliang Liu.
\newblock Alip: Adaptive language-image pre-training with synthetic caption.
\newblock In \emph{Proceedings of the IEEE/CVF International Conference on
  Computer Vision}, pages 2922--2931, 2023.

\bibitem[Young et~al.(2014)Young, Lai, Hodosh, and
  Hockenmaier]{young2014flickr}
Peter Young, Alice Lai, Micah Hodosh, and Julia Hockenmaier.
\newblock From image descriptions to visual denotations: New similarity metrics
  for semantic inference over event descriptions.
\newblock \emph{Transactions of the Association for Computational Linguistics},
  2:\penalty0 67--78, 2014.

\bibitem[Yu et~al.(2022)Yu, Wang, Vasudevan, Yeung, Seyedhosseini, and
  Wu]{yu2022coca}
Jiahui Yu, Zirui Wang, Vijay Vasudevan, Legg Yeung, Mojtaba Seyedhosseini, and
  Yonghui Wu.
\newblock Coca: Contrastive captioners are image-text foundation models.
\newblock \emph{arXiv preprint arXiv:2205.01917}, 2022.

\bibitem[Yue et~al.(2024)Yue, Ni, Zhang, Zheng, Liu, Zhang, Stevens, Jiang,
  Ren, Sun, Wei, Yu, Yuan, Sun, Yin, Zheng, Yang, Liu, Huang, Sun, Su, and
  Chen]{yue2023mmmu}
Xiang Yue, Yuansheng Ni, Kai Zhang, Tianyu Zheng, Ruoqi Liu, Ge~Zhang, Samuel
  Stevens, Dongfu Jiang, Weiming Ren, Yuxuan Sun, Cong Wei, Botao Yu, Ruibin
  Yuan, Renliang Sun, Ming Yin, Boyuan Zheng, Zhenzhu Yang, Yibo Liu, Wenhao
  Huang, Huan Sun, Yu~Su, and Wenhu Chen.
\newblock Mmmu: A massive multi-discipline multimodal understanding and
  reasoning benchmark for expert agi.
\newblock In \emph{Proceedings of CVPR}, 2024.

\bibitem[Zhai et~al.(2023)Zhai, Mustafa, Kolesnikov, and Beyer]{zhai2023siglip}
Xiaohua Zhai, Basil Mustafa, Alexander Kolesnikov, and Lucas Beyer.
\newblock Sigmoid loss for language image pre-training.
\newblock In \emph{Proceedings of the IEEE/CVF international conference on
  computer vision}, pages 11975--11986, 2023.

\bibitem[Zhang et~al.(2022)Zhang, Zhang, Hu, Chen, Li, Dai, Wang, Yuan, Hwang,
  and Gao]{zhang2022glipv2}
Haotian Zhang, Pengchuan Zhang, Xiaowei Hu, Yen-Chun Chen, Liunian~Harold Li,
  Xiyang Dai, Lijuan Wang, Lu~Yuan, Jenq-Neng Hwang, and Jianfeng Gao.
\newblock Glipv2: unifying localization and vl understanding.
\newblock In \emph{36th Conf. Neural Inf. Process. Syst. NeurIPS}, 2022.

\bibitem[Zhang et~al.(2025)Zhang, Li, Zhang, Pu, Cahyono, Hu, Liu, Zhang, Yang,
  Li, et~al.]{zhang2025lmms}
Kaichen Zhang, Bo~Li, Peiyuan Zhang, Fanyi Pu, Joshua~Adrian Cahyono, Kairui
  Hu, Shuai Liu, Yuanhan Zhang, Jingkang Yang, Chunyuan Li, et~al.
\newblock Lmms-eval: Reality check on the evaluation of large multimodal
  models.
\newblock In \emph{Findings of the Association for Computational Linguistics:
  NAACL 2025}, pages 881--916, 2025.

\bibitem[Zheng et~al.(2024)Zheng, Zhang, Wu, Lu, Ma, Jin, Chen, and
  Shen]{zheng2024dreamlip}
Kecheng Zheng, Yifei Zhang, Wei Wu, Fan Lu, Shuailei Ma, Xin Jin, Wei Chen, and
  Yujun Shen.
\newblock Dreamlip: Language-image pre-training with long captions.
\newblock In \emph{European Conference on Computer Vision}, pages 73--90.
  Springer, 2024.

\bibitem[Zhou et~al.(2017)Zhou, Zhao, Puig, Fidler, Barriuso, and
  Torralba]{zhou2017scene}
Bolei Zhou, Hang Zhao, Xavier Puig, Sanja Fidler, Adela Barriuso, and Antonio
  Torralba.
\newblock Scene parsing through ade20k dataset.
\newblock In \emph{Proceedings of the IEEE conference on computer vision and
  pattern recognition}, pages 633--641, 2017.

\bibitem[Zhu et~al.(2024)Zhu, Chen, Shen, Li, and Elhoseiny]{zhu2024minigpt}
Deyao Zhu, Jun Chen, Xiaoqian Shen, Xiang Li, and Mohamed Elhoseiny.
\newblock Minigpt-4: Enhancing vision-language understanding with advanced
  large language models.
\newblock In \emph{The Twelfth International Conference on Learning
  Representations}, 2024.

\end{thebibliography}

\clearpage

\beginappendix

\section{Zero-shot Retrieval}
\label{sec:retrieval}
We evaluate GenLIP on zero-shot image-text retrieval with the Flickr30K~\cite{young2014flickr} dataset. Following VisualGPTScore~\cite{linrevisiting}, we rank image-text pairs with generative scores without training a task-specific retrieval head. Table~\ref{tab:retrieval} compares contrastive baselines in the upper block with generative models in the lower block. The results show that retrieval remains challenging for generative objectives relative to contrastive image-text discrimination.

\begin{table}[!htbp]
  \caption{Zero-shot retrieval on Flickr30K. We report Recall@1 for text-to-image (T$\rightarrow$I) and image-to-text (I$\rightarrow$T) retrieval.}
  \label{tab:retrieval}
  \centering
  \small
  \begin{tabular}{lccc}
    \toprule
    Method & Arch & T$\rightarrow$I & I$\rightarrow$T \\
    \midrule
    CLIP & L/14 & 66.9 & 87.7 \\
    SigLIP & So/14 & 75.3 & 91.0 \\
    SigLIP2 & So/16 & 85.3 & 95.9 \\
    \midrule
    IG Captioner & L/14 & 76.2 & 64.7 \\
    GenLIP & L/16 & 68.7 & 52.6 \\
    GenLIP & So/16 & 69.2 & 57.0 \\
    GenLIP & g/16 & 72.5 & 63.7 \\
    \bottomrule
  \end{tabular}
\end{table}

\FloatBarrier

\section{Cambrian-1 style Evaluation}
\label{sec:cambrian-1_style}
Based on the frozen evaluation in Table~\ref{tab:Qwen2_5-1_5B} and Table~\ref{tab:Qwen2_5-7B}, we also organize our evaluation results in a Cambrian-1-style~\cite{tong2024cambrian} benchmark suite by adding benchmarks such as MMMU~\cite{yue2023mmmu}, MathVista~\cite{lu2024mathvista}, MMVP~\cite{tong2024eyes}, and RealWorldQA\cite{realworldqa2024}. Detailed results are shown in Table~\ref{tab:cambrian_qwen2_5_1_5B} and Table~\ref{tab:cambrian_qwen2_5_7B}. On this extended evaluation suite, GenLIP still shows its performance advantage against strong baselines on all model sizes.

The main experiment section reports the Cambrian-1-style evaluation under the LLaVA-NeXT-Qwen2.5-1.5B setting in Table~\ref{tab:cambrian_qwen2_5_1_5B}. Here, we provide the complementary Qwen2.5-7B results to examine whether the same trend holds with a stronger language backbone. The benchmark grouping, visual-token budget, SFT data, and evaluation protocol are kept the same as in the main text; only the LLM size changes.

As shown in Table~\ref{tab:cambrian_qwen2_5_7B}, GenLIP remains consistently strong under the larger LLM. At L/16, GenLIP achieves the best overall average among the compared L-scale encoders. At So/16 and g/16, GenLIP improves the average score over SigLIP2 by $1.6$ and $3.0$ points, respectively. The gains are especially clear on OCR and chart-oriented benchmarks such as ChartQA, OCRBench, TextVQA, and DocVQA, consistent with the main-text observation that GenLIP's generative pretraining is particularly effective for fine-grained visual-text alignment.

\begin{table}[!htbp]
  \caption{Cambrian-1-style frozen visual representation evaluation under LLaVA-NeXT-Qwen2.5-7B. Except for the LLM size, all settings are the same as those used in the Qwen2.5-1.5B setting.}
  \label{tab:cambrian_qwen2_5_7B}
  \centering
  \resizebox{\linewidth}{!}{
  \begin{tabular}{lcc | cccc  cccc  cccc  cccc  c}
    \toprule
    \multirow{2}{*}{\B{Model}} & \multirow{2}{*}{\B{Arch}} & \multirow{2}{*}{\B{Data}} & \multicolumn{4}{c}{\B{General VQA}} & \multicolumn{4}{c}{\B{Knowledge}}  & \multicolumn{4}{c}{\B{OCR \& Chart}} & \multicolumn{4}{c}{\B{Vision-Centric}} \\
    \cmidrule(lr){4-7} \cmidrule(lr){8-11} \cmidrule(lr){12-15} \cmidrule(lr){16-19}
    & &
    & \rotatebox{90}{\scriptsize{MME-P}}
    & \rotatebox{90}{\scriptsize{MMB}}
    & \rotatebox{90}{\scriptsize{SEED-I}}
    & \rotatebox{90}{\scriptsize{GQA}}
    & \rotatebox{90}{\scriptsize{SQA}}
    & \rotatebox{90}{\scriptsize{MMMU}}
    & \rotatebox{90}{\scriptsize{MathV}}
    & \rotatebox{90}{\scriptsize{AI2D}}
    & \rotatebox{90}{\scriptsize{ChartQA}}
    & \rotatebox{90}{\scriptsize{OCR-B}}
    & \rotatebox{90}{\scriptsize{TextVQA}}
    & \rotatebox{90}{\scriptsize{DocVQA}}
    & \rotatebox{90}{\scriptsize{MMVP}}
    & \rotatebox{90}{\scriptsize{RWQA}}
    & \rotatebox{90}{\scriptsize{CV-2D}}
    & \rotatebox{90}{\scriptsize{CV-3D}}
    & \multirow[t]{2}{*}{\rotatebox{90}{AVG}} \\
    \midrule
    CLIP~\cite{radford2021clip} & L/14 & 12.8B & 1316 & 73.6 & 69.9 & 39.6 & 85.2 & 46.4 & 48.3 & 76.3 & 36.6 & 29.6 & 52.6 & 48.4 & 36.0 & 58.0 & 60.4 & \B{65.3} & 55.7 \\
    AIMv2~\cite{fini2025aimv2} & L/14 & 12.0B & 1240 & 74.2 & 70.9 & 37.9 & 85.2 & 45.9 & 48.0 & 76.9 & 36.8 & 30.9 & 54.5 & 46.6 & 48.7 & 56.9 & 63.0 & 63.3 & 56.4 \\
    OVision2~\cite{liu2025openvision2} & L/16 & 12.8B & \B{1325} & 74.5 & 71.1 & 47.2 & 85.9 & 45.4 & 45.9 & 78.4 & 42.5 & 49.9 & 58.8 & 49.5 & 42.0 & \B{59.1} & 62.5 & 64.2 & 58.9 \\
    SigLIP~\cite{zhai2023siglip} & L/16 & 40.0B & 1275 & \B{76.2} & 71.6 & 46.2 & \B{86.7} & \B{48.3} & 52.0 & 79.3 & 41.7 & 45.7 & 56.0 & 50.5 & 46.7 & 58.7 & 63.0 & 59.3 & 59.1 \\
    \rowcolor{lightcyan}
    GenLIP & L/16 & 8.0B & 1320 & 74.3 & \B{73.3} & \B{51.3} & 85.4 & 44.8 & \B{53.6} & \B{80.4} & \B{52.7} & \B{59.2} & \B{62.9} & \B{61.7} & \B{50.0} & 58.0 & \B{67.9} & 60.3 & \B{62.6} \\
    \midrule
    SigLIP2~\cite{tschannen2025siglip2} & So/16 & 40.0B & 1422 & \B{77.7} & 72.6 & 52.2 & \B{87.1} & \B{47.2} & 53.7 & \B{81.3} & 46.6 & 55.6 & 63.5 & 56.3 & 46.0 & \B{60.4} & 66.2 & \B{66.3} & 62.7 \\
    \rowcolor{lightcyan}
    GenLIP & So/16 & 8.0B & \B{1424} & 76.6 & \B{73.1} & \B{52.4} & 86.4 & 46.6 & \B{55.2} & 81.0 & \B{55.3} & \B{63.5} & \B{65.7} & \B{66.3} & \B{48.0} & 59.1 & \B{66.4} & 61.8 & \B{64.3} \\
    \midrule
    SigLIP2~\cite{tschannen2025siglip2} & g/16 & 40.0B & 1422 & \B{78.0} & 72.8 & 49.3 & \B{87.7} & \B{49.0} & 53.6 & \B{81.0} & 47.2 & 55.6 & 63.5 & 56.3 & 50.7 & 60.3 & \B{65.8} & 65.6 & 63.0 \\
    \rowcolor{lightcyan}
    GenLIP & g/16 & 8.0B & \B{1483} & 77.3 & \B{73.6} & \B{54.5} & 87.0 & 47.6 & \B{55.8} & \B{81.0} & \B{57.1} & \B{65.9} & \B{66.8} & \B{69.0} & \B{52.0} & \B{62.1} & 65.4 & \B{67.4} & \B{66.0} \\
    \bottomrule
  \end{tabular}
  }
\end{table}

\FloatBarrier

\section{Supplementary Qualitative Results}
We provide additional qualitative results that complement the ``Let ViT Speak'' analysis in Sec.~\ref{subsec:4_0vit_speak}. Besides illustrating the strengths of GenLIP, these cases also expose its remaining failure modes on challenging detail-sensitive inputs.

\definecolor{l16s1}{HTML}{528D67}
\definecolor{so16s1}{HTML}{215F9A}
\definecolor{g16s1}{HTML}{7C5FA3}
\definecolor{l16s2}{HTML}{417654}
\definecolor{so16s2}{HTML}{163E64}
\definecolor{g16s2}{HTML}{78548D}

\begin{figure}[!htbp]
  \centering
  \includegraphics[width=0.96\linewidth]{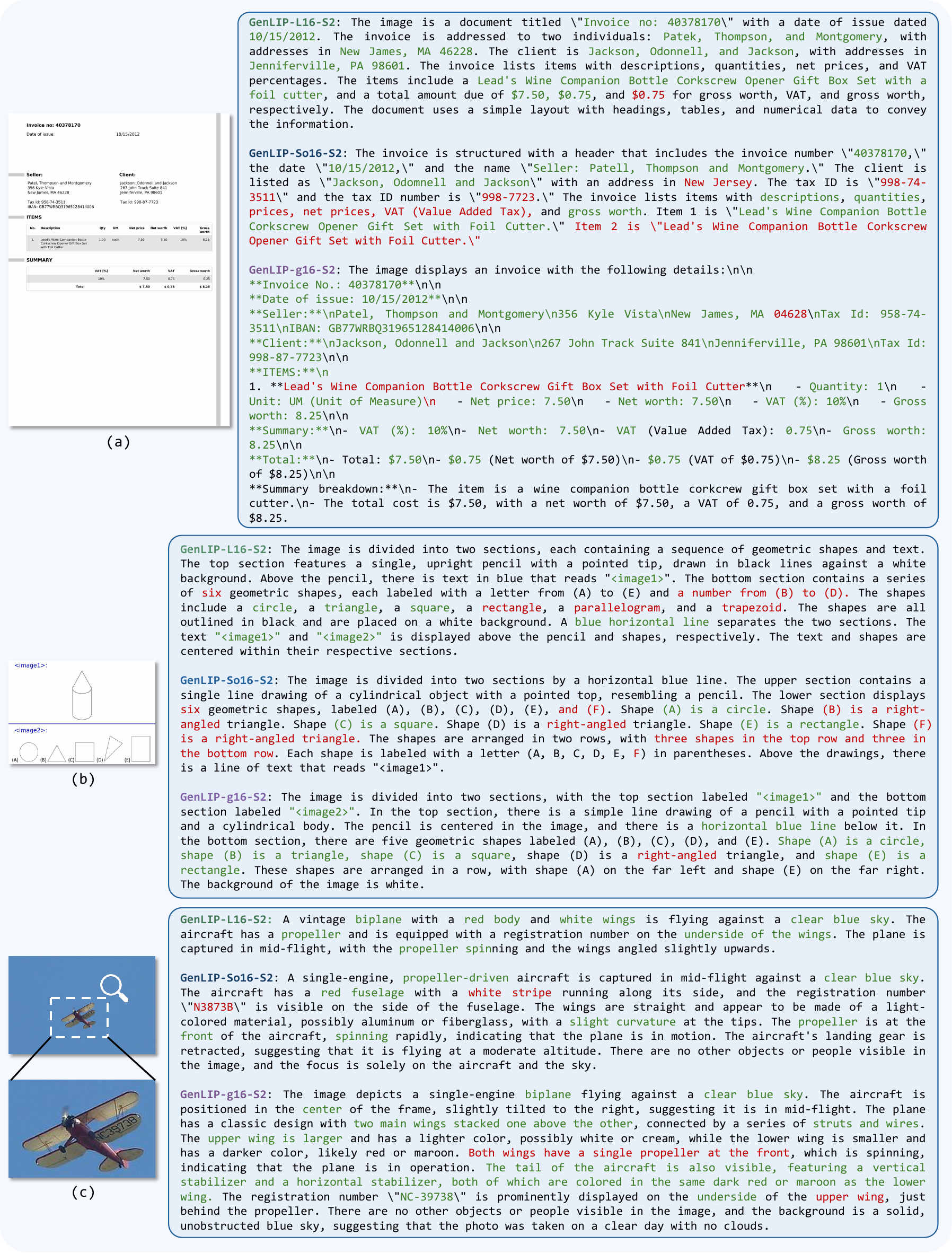}
  \caption{\textbf{Additional OCR Cases.} Representative GenLIP generations on three challenging examples that require fine-grained detail recognition.}
  \label{fig:caption-2}
\end{figure}

In Figure~\ref{fig:caption-2}, we further evaluate GenLIP on challenging OCR-heavy examples. These three cases test (a) receipt understanding, (b) geometric-shape counting and placement, and (c) recognition of tiny characters and numbers. All three model variants show non-trivial OCR ability, although clear errors remain:

\noindent\textbf{(a)} In the first case (Figure~\ref{fig:caption-2}(a)), \textcolor{l16s2}{GenLIP-L16-S2} recognizes most characters but fails on the long number sequences (Tax Id and IBAN) and the two tables. \textcolor{so16s2}{GenLIP-So16-S2} encounters similar difficulties and produces repeated output. In contrast, \textcolor{g16s2}{GenLIP-g16-S2} recovers the table structure much more accurately, missing only one number and the word ``Opener''.

\noindent\textbf{(b)} In the second case (Figure~\ref{fig:caption-2}(b)), \textcolor{l16s2}{GenLIP-L16-S2} and \textcolor{so16s2}{GenLIP-So16-S2} make mistakes in both the number and placement of geometric shapes. \textcolor{g16s2}{GenLIP-g16-S2} is substantially more accurate, with the main remaining error being that it identifies the acute triangle in the bottom row as a right triangle.

\noindent\textbf{(c)} In the last case (Figure~\ref{fig:caption-2}(c)), \textcolor{l16s2}{GenLIP-L16-S2} fails to detect the number on the plane, and \textcolor{so16s2}{GenLIP-So16-S2} outputs the wrong number. \textcolor{g16s2}{GenLIP-g16-S2} identifies the number correctly but still makes a spatial error.

Overall, these examples show that GenLIP already acquires meaningful OCR ability even without an OCR-specific pretraining corpus. This ability scales clearly with model size: larger models recognize and describe subtle details more accurately. At the same time, the observed errors show that long number strings, precise spatial layouts, and tiny text remain challenging. These cases help explain both the strong Doc\&OCR performance of GenLIP and the residual gaps that remain in detail-sensitive settings.

\begin{figure}[H]
  \centering
  \includegraphics[width=\linewidth]{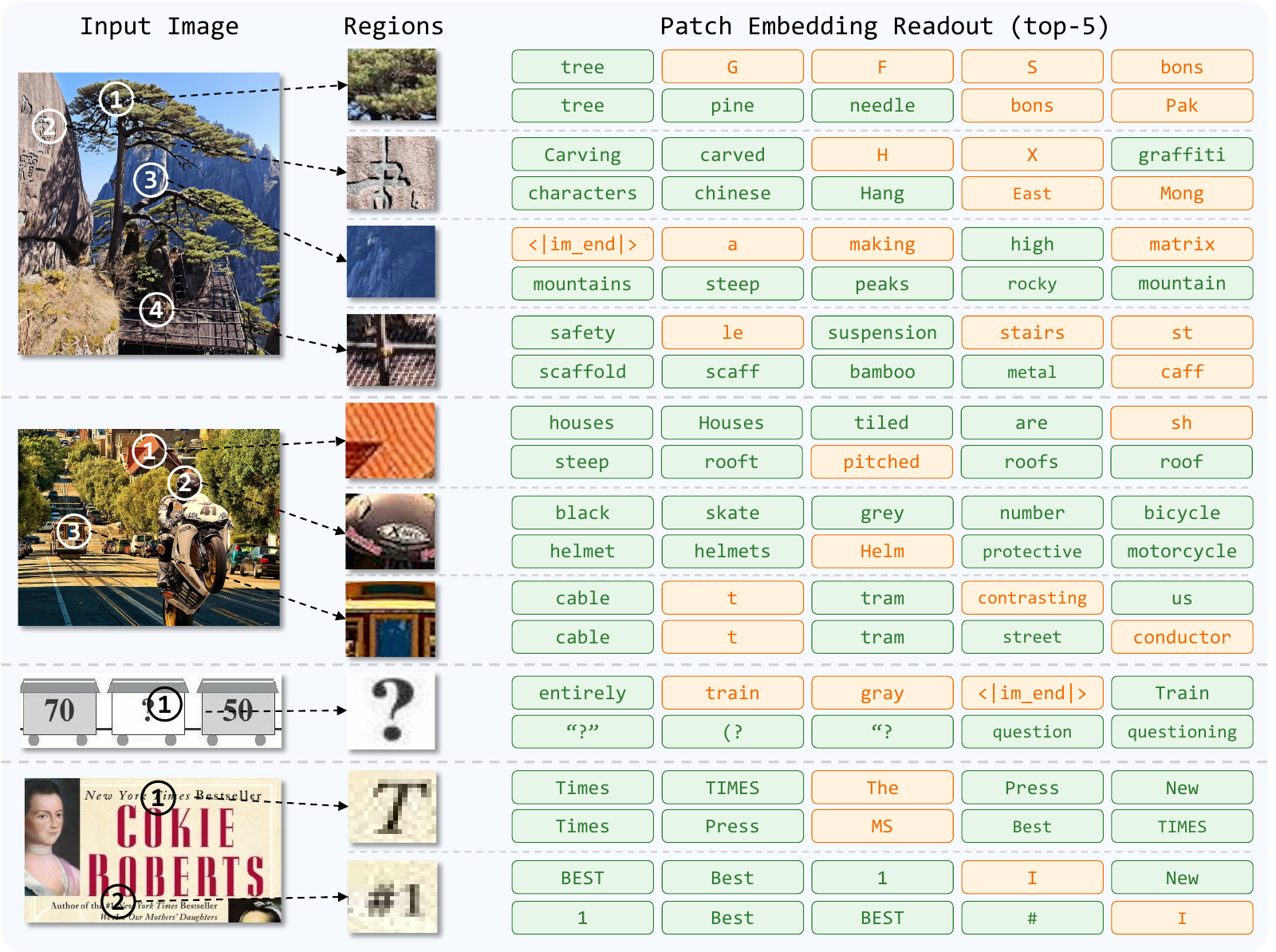}
  \caption{\textbf{Additional Patch Semantics Cases.} Further examples of direct semantic readout from image patch embeddings for \textcolor{g16s1}{GenLIP-g16-S1} and \textcolor{g16s2}{GenLIP-g16-S2}. The stage-2 model generally shows stronger alignment.}
  \label{fig:patchsemantics-2}
\end{figure}

In Figure~\ref{fig:patchsemantics-2}, we provide four more cases in addition to Figure~\ref{fig:patchsemantics-1} using the same model configurations.

\section{Additional Implementation Details}
\paragraph{\textbf{Frozen Visual Representation Evaluation}.} We summarize the training settings for frozen visual representation evaluation. Relative to the default LLaVA-NeXT~\cite{liu2024llavanext} setup, we make three modifications: (i) we replace the original LLM LLaMA3-8B with Qwen2.5 models; (ii) we replace the original 780K SFT dataset with the 3M SFT dataset from LLaVA-OneVision; and (iii) we use the simplest image preprocessing pipeline, consisting only of resize and crop operations, without ``anyres'' processing designed for high-resolution images. All other training settings remain unchanged, including the optimization hyperparameters, batch size, iterations, and the 2-layer MLP projector.

\paragraph{\textbf{Metric Aggregation}.} For the frozen visual representation results in Sec.~\ref{subsec:4_1eval}, we report \emph{ALL AVG} as the unweighted mean over all 14 benchmarks. Because MME-P is reported on a 0--2000 scale, we divide it by 20 to map it into the range $[0, 100]$ before averaging. We keep the CIDEr scores on caption benchmarks unchanged in the \emph{ALL AVG} calculation.

\paragraph{\textbf{Pretraining Implementation}.} The main hyperparameters of GenLIP pretraining are summarized in Table~\ref{tab:hyperparameters}. Stage 1 uses fixed $224\times224$ inputs and trains for 8B samples from Recap-DataComp-1B to learn strong foundational visual representations. Stage 2 then adapts the model for one epoch on the 39M samples from BLIP3o-Long-Caption, Infinity-MM (stage1), and CapRL with native aspect ratios, resizing each image so that the number of visual tokens stays within $[16, 1024]$. For efficiency, we pack variable-length samples into sequences with a maximum length of $16{,}384$ tokens and implement exact per-sample Prefix-LM masking with PyTorch FlexAttention. Because the second stage contains much longer sequences on average, its global batch size is reduced to 3.6K, and its peak learning rate is reduced to $1e\!-4$, as detailed in Sec.~\ref{sec:methods}. The remaining optimization settings follow Stage 1.

\section{Discussion: Attention Sink and Gated Attention}
In GenLIP, we observe the ``attention sink'' phenomenon, which has also been reported in prior transformer studies in both vision~\cite{darcet2023registers} and language~\cite{xiao2023efficient, qiu2025gated}. At a high level, attention sink arises from the sum-to-one normalization of softmax attention: for each query token, the model must distribute a fixed unit mass over all keys. In practice, this often encourages the network to allocate a disproportionate amount of attention to a small subset of tokens that behave like persistent ``registers'' and absorb information from many other positions.

The manifestation of attention sink depends on the attention pattern of the modality. In vision transformers with bidirectional self-attention, sink behavior often appears as a small number of tokens in low-semantic regions that attract attention from many other visual tokens~\cite{darcet2023registers} and exhibit unusually high norm. In contrast, in autoregressive language models the phenomenon is typically more structured: early tokens, especially the first token, tend to receive disproportionately large attention weights from subsequent positions regardless of content. As discussed in StreamingLLM~\cite{xiao2023efficient}, such sink tokens may preserve useful global context information and can even be exploited for efficient long-context inference. This difference is largely explained by the underlying attention mechanism: full attention in vision does not privilege a fixed position a priori, whereas causal attention in language naturally makes early tokens accessible to all later tokens and therefore encourages early ones to serve as shared context carriers.

The Prefix-LM attention used in GenLIP combines bidirectional attention over the visual prefix with causal attention over the text suffix, making its sink behavior closer to that of autoregressive language models. The input sequence follows the organization $[v_0, \ldots, v_M, t_0, \ldots, t_L]$, positioning visual tokens as the prefix for text generation. Because the loss is applied only to text-token predictions, the model tends to compress information useful for generation into a few preceding visual tokens that are broadly accessible to the text tokens. Under this structure, the first visual token $v_0$ becomes a particularly favorable sink candidate, since it can be attended by all subsequent text tokens and thus can act as a compact carrier of global visual context.

Empirically, we find that this behavior can partially degrade the discriminative quality of the visual representation, as reflected by the degraded attentive-probing results of the ``w/o GA'' variant in Table~\ref{tab:frozen_dis}. This observation motivates the introduction of gated attention in GenLIP, which alleviates overly concentrated sink behavior and improves the quality of the learned visual features. We also note that many encoder-decoder generative VLP architectures are less affected by this issue. Because the visual encoder and text decoder are separated, sink behavior is largely confined to the decoder side and therefore has much weaker direct impact on the quality of the visual encoder representations.

\section{Discussion: GenLIP Meets Language Priors}
We observe a counterintuitive result in Table~\ref{tab:sail_comparison}: initializing the single-transformer vision encoder from a pretrained Qwen3-0.6B language model does not lead to stronger frozen visual representations in our setting.
Adding gated attention to the Qwen-initialized SAIL variant improves the overall average only modestly, from $53.6$ to $54.8$, whereas training the same SAIL-style architecture from scratch reaches $56.0$, close to GenLIP-So/16 at $56.3$.
Since these variants use the same 1B-sample pretraining data and similar model scale, this result suggests that the gap is unlikely to be explained by model capacity or the single-transformer architecture itself.

A plausible explanation is that strong LLM initialization changes the optimization path of caption-based vision-language pretraining.
With a pretrained language model, the model already has substantial next-token prediction ability on the language side.
As a result, the captioning objective can be partially satisfied by adapting this strong language prior with relatively weak visual conditioning, rather than by establishing sufficiently dense vision-to-text information transfer.
In other words, the training process may become closer to adapting a language model into a visually conditioned model, instead of learning a unified multimodal representation from scratch.
This bias can be undesirable for our goal, because the final model is used as a standalone vision encoder whose quality depends on distributed and grounded visual representations.

\begin{figure*}[htbp]
    \centering
    \includegraphics[width=1.0\linewidth]{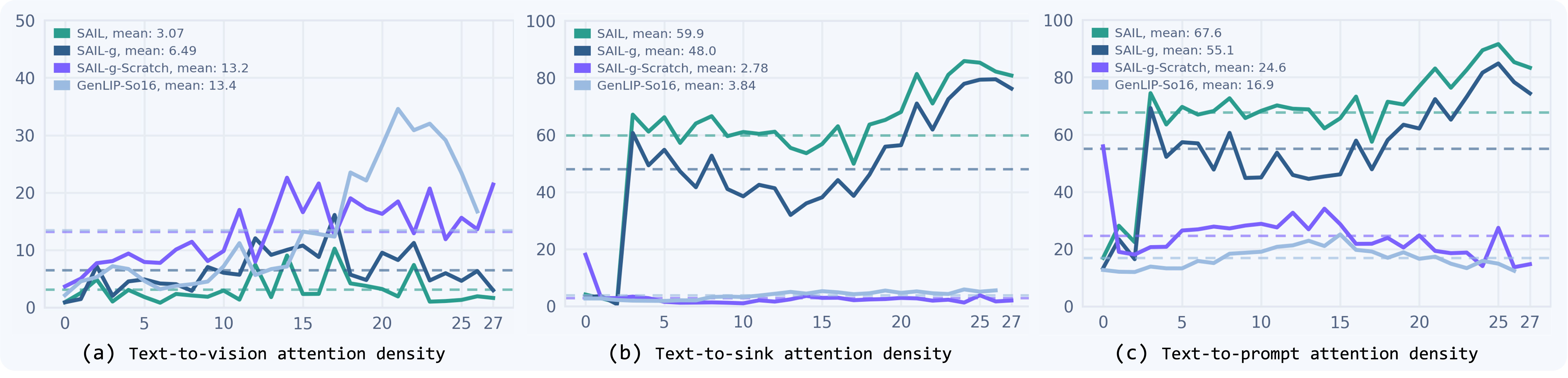}
    \caption{\textbf{Layer-wise attention allocation across token groups}. We analyze the attention allocation of generated text tokens over different target token groups with models in Table~\ref{tab:sail_comparison}. We report the attention density from generated text tokens to (a) vision tokens, (b) sink tokens (the first two prompt tokens), and (c) prompt tokens. Dashed lines denote the layer-averaged attention densities. The X-axis corresponds to the layer index. Unlike the original SAIL implementation, our implementation does not insert the special tokens `<vision>' and `</vision>' around the visual patch tokens.
    }
    \label{fig:sail_attn_alloc}
\end{figure*}

\begin{figure*}[htbp]
    \centering
    \includegraphics[width=1.0\linewidth]{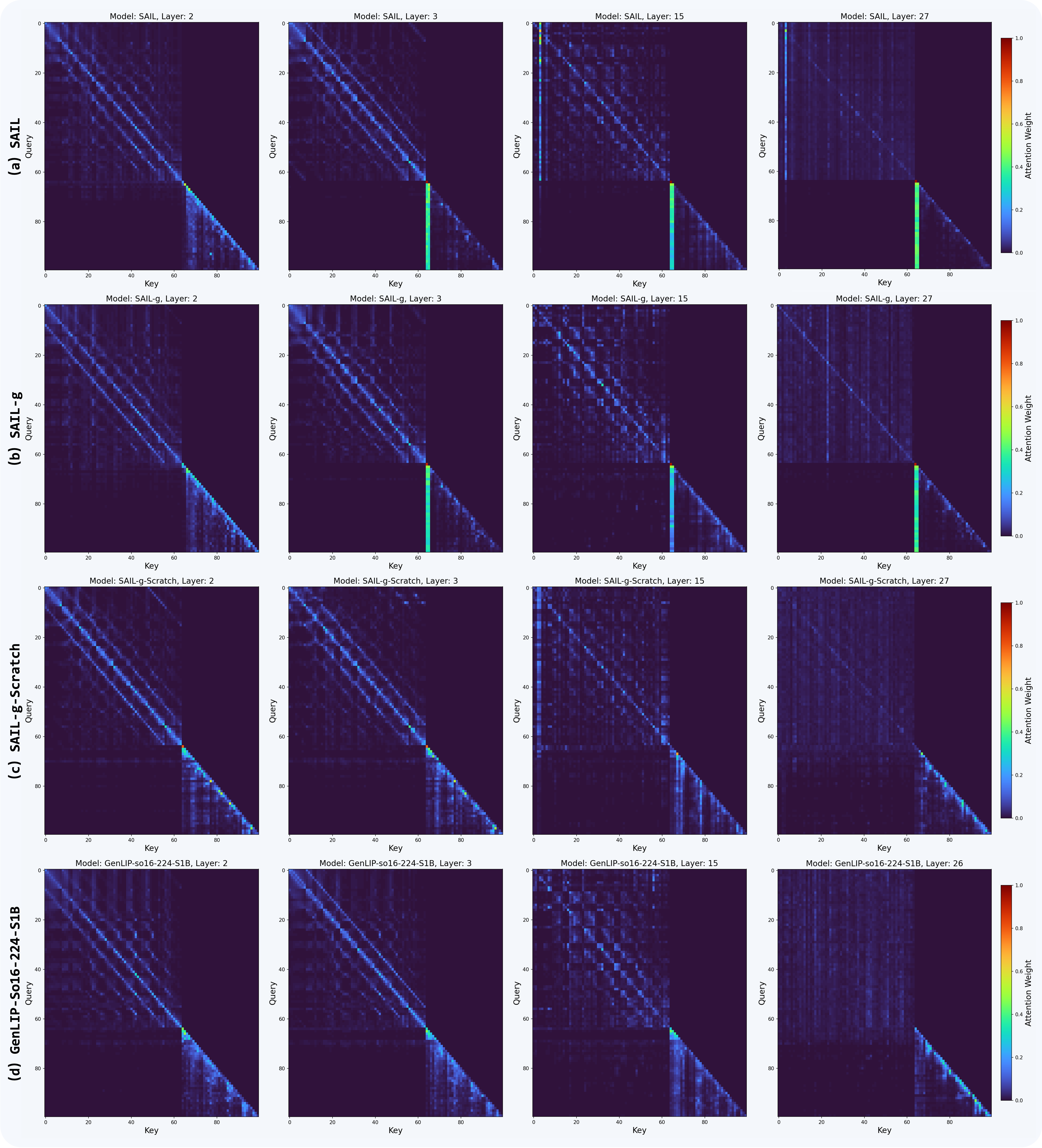}
    \caption{Layer-wise attention maps of controlled SAIL-style model variants.
    We visualize the head-averaged attention maps of four model variants in Table~\ref{tab:sail_comparison}: (a) SAIL with Qwen-0.6B initialization, (b) SAIL with gated attention, (c) SAIL with gated attention trained from scratch, and (d) a controlled GenLIP-So16 model.  From left to right, we show the attention maps from the 2nd, 3rd, 15th, and final layers of each model. Each map is averaged over all attention heads in the corresponding layer, with rows denoting query tokens and columns denoting key tokens. The input image is resized to 128×128, yielding the first 64 visual tokens, followed by 6 prompt tokens and the first 30 generated text tokens. The two Qwen-initialized SAIL variants, (a) and (b), exhibit clear attention-sink behavior in the text tokens, where generated text tokens assign disproportionately high attention to the first two prompt tokens.}
    \label{fig:attn_map}
\end{figure*}

This interpretation is consistent with the attention patterns in Figures~\ref{fig:attn_map} and~\ref{fig:sail_attn_alloc}.
After modeling both vision and text modalities in early layers, the Qwen-initialized SAIL variants allocate much of the generated-text attention to language-side anchors, including the first two prompt token (sink tokens) and all prompt tokens, while assigning limited attention mass to the visual prefix.
Gated attention reduces this concentration but does not fully remove the inherited language-side dependency under LLM initialization.
In contrast, when the same SAIL architecture is trained from scratch, the model no longer has a strong language shortcut and relies more directly on visual evidence to predict captions.
Consequently, the SAIL-g-Scratch variant shifts much more attention toward visual tokens, and its modeling pattern becomes close to that of GenLIP-So16.

The attention patterns in Figures~\ref{fig:attn_map} and~\ref{fig:sail_attn_alloc} suggest that removing language initialization substantially increases the pressure to ground language prediction in the visual prefix.
In this controlled vision-encoder pretraining setting, strong language initialization appears to introduce an initialization-induced attention allocation bias: it preserves strong language modeling ability, but can reduce the optimization pressure to learn distributed visual grounding from image-caption data.
For pretraining a modular vision encoder within a single-transformer vision-language pretraining framework, training from scratch therefore better aligns the captioning objective with the goal of learning grounded visual representations.

\end{document}